\documentclass{article}
\pdfoutput=1
 \UseRawInputEncoding
\PassOptionsToPackage{square,comma}{natbib}

\usepackage{wrapfig}
\usepackage[preprint]{neurips_2024}

\usepackage[utf8]{inputenc} 
\usepackage[T1]{fontenc}    
\usepackage{hyperref}       
\usepackage{url}            
\usepackage{booktabs}       
\usepackage{amsfonts}       
\usepackage{nicefrac}       
\usepackage{microtype}      
\usepackage{xcolor}         
\usepackage{graphicx}
\usepackage{amsmath}
\usepackage{algorithmic}
\usepackage{algorithm}
\usepackage{enumitem}
\usepackage{booktabs}
\usepackage{pifont}
\usepackage{subcaption}
\usepackage{authblk}
\usepackage[title]{appendix}

\title{Domain Generalization Guided by Large-Scale Pre-Trained Priors}

\author[1]{\textbf{Zongbin Wang}}
\author[1]{\textbf{Bin Pan}}
\author[1]{\textbf{Shiyu Shen}}
\author[2]{\textbf{Tianyang Shi}}
\author[3]{\textbf{Zhenwei Shi}} 

\affil[1]{Nankai University}
\affil[2]{ByteDance}
\affil[2]{Beihang University}

\begin{document}

	\maketitle

	\begin{abstract}

Domain generalization (DG) aims to train a model from limited source domains, allowing it to generalize to unknown target domains. Typically, DG models only employ large-scale pre-trained models during the initialization of fine-tuning. However, large-scale pre-trained models already possess the ability to resist domain shift. If we reference pre-trained models continuously during fine-tuning to maintain this ability, it could further enhance the generalization ability of the DG model. For this purpose, we introduce a new method called Fine-Tune with Large-scale pre-trained Priors (FT-LP), which incorporates the pre-trained model as a prior into the DG fine-tuning process, ensuring that the model refers to its pre-trained model at each optimization step. FT-LP comprises a theoretical framework and a simple implementation strategy. In theory, we verify the rationality of FT-LP by introducing a generalization error bound with the pre-trained priors for DG. In implementation, we utilize an encoder to simulate the model distribution, enabling the use of FT-LP when only pre-trained weights are available. In summary, we offer a new fine-tuning method for DG algorithms to utilize pre-trained models throughout the fine-tuning process. Through experiments on various datasets and DG models, our proposed method exhibits significant improvements, indicating its effectiveness.

    
	\end{abstract}
	
	\section{Introduction}
    In the current landscape of machine learning, the majority of models hinge upon a simple assumption: the training source and testing target data follow the same distributions \cite{ref1}. However, reality may diverge from this assumption, rendering these models less robust when confronted with significant shifts between the source and target domains \cite{ref2}. Examples of such shifts include variations in data between sunny and rainy weather, changes in texture, background differences, and alterations in style \cite{ref3,ref4,ref5,ref6}. Domain generalization (DG) aims to address these challenges, helping models to reduce prediction loss in both known source and unknown target domains simultaneously  \cite{ref7}.

	      \begin{figure*}[htbp]
		\centering
		\includegraphics[width=1\textwidth]{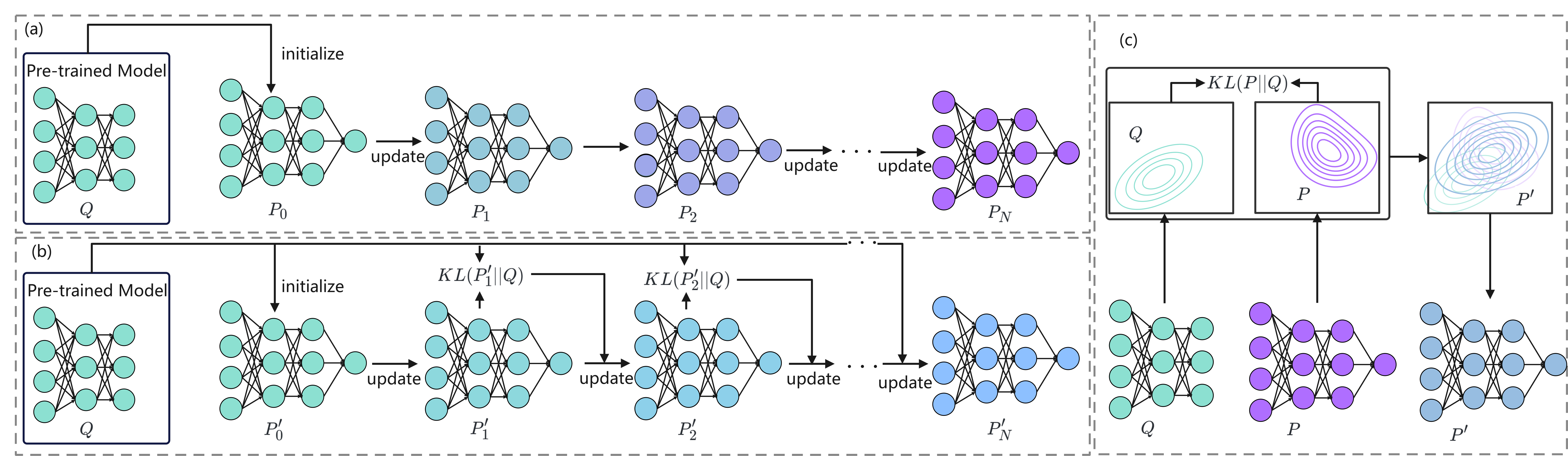}
		\caption{
			Figure (a) shows the standard DG fine-tuning algorithm, where the pre-trained model $Q$ is directly fine-tuned to $P_N$. Figure (b) shows our method, which includes $Q$ at every step of fine-tuning. Figure (c) shows the details: we regard $Q$ as the prior for $P$ and adjust $P$ based on the discrepancy between $P$ and $Q$, resulting in $P'$ for further fine-tuning. }
		\label{alg}
	\end{figure*} 
	The implementation of DG methods often begins by initializing the training models with large-scale pre-trained models, followed by fine-tuning using corresponding algorithms \cite{ref2,ref8,ref9,ref10,ref54,domainbed}. This fine-tuning method indicates that the DG methods utilize information from pre-trained models to improve generalization \cite{ref2}. However, if the pre-trained model is not consistently referenced during optimization, there may be incomplete utilization of this information. This could be significant because research \cite{ref14,ref17,ref18} has shown that pre-trained models already possess abilities to resist domain shifts, while suboptimal fine-tuning may weaken this capability rather than enhance it. Therefore, we argue that in DG problems, it is important for the training model to keep referencing its pre-trained model during fine-tuning. This may help the model improve the ability to resist domain shifts from the pre-trained model, which may not be easily achievable through DG methods alone \cite{domainbed,ref59}. 

	However, integrating pre-trained models into the fine-tuning of DG poses significant challenges, because DG methods have varying objectives \cite{ref2,ref15,SAGM,DANN}. The data-dependent prior presents a promising approach   \cite{ref42,ref41,ref23}. It is constructed within the Probabilistic Approximation Complexity-Bayesian (PAC-Bayes) bound \cite{ref40,ref20}, allowing for the integration of another model as a prior to guide training model, thereby reducing the generalization error. However, both PAC-Bayes bound and data-dependent prior merely consider the situation of DG settings \cite{ref15}, and their focus lies in model distribution rather than model weights \cite{ref23,ref20}. Under these limitations, it is challenging to directly apply them to DG problems.

	Therefore, we propose a new fine-tuning objective, Fine-Tune with Large-scale pre-trained Priors (FT-LP), which re-formulates the DG objective to ensure that the model references its pre-trained models at each step of optimization. As Fig.\ref{alg} shows, when fine-tuning the DG model, FT-LP additionally adjusts the training model distribution by controlling its discrepancy from the pre-trained prior. This ensures that the FT-LP remains compatible with DG methods, preserving their assumptions and internal structure. In theory, we introduce a new PAC-Bayes bound that measures the generalization error between the source and target domains, and we prove that FT-LP can reduce this error. In implementation, we encode pre-trained model weights to simulate the distribution of priors, thereby effectively applying FT-LP  to the DG methods. We employ various DG algorithms on Domainbed \cite{domainbed} to evaluate the effectiveness of FT-LP, confirming its efficacy.

	Overall, our research has made several contributions:
	
	\begin{enumerate}
	     \item[1.] We develop a new fine-tuning method that incorporates pre-trained models as priors in the Domain Generalization process, ensuring that training models refer to pre-trained models at every optimization step.
	    \item[2.] We propose a new generalization error upper bound for DG problems and show that using pre-trained models as priors into the bound can reduce this error.
	    \item[3.] We propose a strategy to apply our method to DG algorithms by simulating prior distributions, making the training models continually reference their pre-trained models.
	    
		
	\end{enumerate}
	\section{Related Work}
	\textbf{Domain Generalization.}  DG aims to train a model across multiple known source domains to enable it to generalize to unknown target domains \cite{dgx}.  This concept encompasses various theories, resisting domain shifts from different perspectives \cite{ref2,ref8}. Some methods minimize the feature distance between training domains to discover domain-invariant features \cite{DANN,CDANN,MMD,CORAL,ref55}. Others use data augmentation to expose the model to a broader range of distributions, enhancing its robustness \cite{ref9,SagNet,Mixup,ref57}. Methods in robust learning suggest minimizing source domain worst-case loss when we know the source-target domain distance \cite{DRO} or employing meta-learning to simulate domain shifts to improve generalization capability \cite{MLDG}. Some alter DG optimization goals by assuming specific invariant feature distributions \cite{ref16,ref15,RIDG}. Additionally, gradient-based methods advocate for maximizing gradient similarity across domains \cite{Fishr}, while others suggest that smoother loss surfaces lead to stronger generalization ability \cite{SAGM}. In this paper, we show theoretically that including pre-trained models in DG  objectives will help improve their generalization abilities.

	 \textbf{Data-dependent priors.} In PAC learning theory, PAC-Bayes bounds serve as vital metrics for assessing the discrepancy between empirical and generalization errors of the models \cite{ref30}. PAC-Bayes bounds often involve the  Kullback-Leibler (KL) divergence term, which measures the difference between the model distribution and a prior independent of training data\cite{ref40,ref20}. The data-dependent prior suggests that utilizing available data can help identify priors that reduce the KL term, thus obtaining a tighter bound \cite{ref42}. This method was first proposed by \cite{ref41}. After that, various strategies have been explored to optimize the KL term, including splitting training data to train priors \cite{ref23,ref52}, using all data priors \cite{ref24}, incorporating data into the distribution \cite{ref46,ref25}, retraining on datasets of pre-trained models \cite{ref45}, and direct training the Bayesian networks \cite{ref53}. However, these methods seldom consider DG settings and face challenges due to Bayesian model optimization issues \cite{ref25}. We propose imposing constraints on target and source losses to establish a new upper bound for DG problems while also integrating large-scale pre-trained models as a prior to obtain a tighter bound.

	\section{Fine-Tuning with Large-scaled pre-trained Priors}
	
	\subsection{Problem setting}

	 We set $S$ as the source domains, $T$ as the target domains, $h\in\mathcal{H}$ is the model which only contains weights, different DG methods may have distinct optimization goals, denoted as $L_{DG}(h(x),y)$. The DG objective can be treated as:
	  \begin{equation}
	  	\mathop{\arg\min}_{h\in\mathcal{H}} E_{\{x,y\} \sim S}L_{DG}(h(x),y). \label{common}
	  \end{equation}
 Typically, these methods initialize $h$ using a pre-trained model $h_0 \in\mathcal{H}$ on a large-scale dataset $D_0$ for fine-tuning. However, as (\ref{common}) shows, they do not consider $h_0$ during fine-tuning, which suggests that the information in $h_0$ might not be fully used. A strategy of incorporating another model into the optimization is through the PAC-Bayes bound \cite{ref40} and data-dependent priors \cite{ref42}. Set the task loss as $R_{S}(h) = E_{\{x,y\} \sim S}R(h(x),y) $, and its empirical form $R_{S_m}(h) = \frac{1}{|S_m|}\sum_{i=1}^{|S_m|} R((h(x_i),y_i)), S_m\subset S, |S_m|=m $, we introduce one bound that will be utilized:
     
(Linear PAC-Bayes bound)  $\forall \beta,\sigma \in (0,1)$, $n \in \mathbb{N}$, $P \in \mathcal{M}_1(\mathcal{H})$, $S_m \subset S$. Define $KL(P||Q)=\int P(h)ln\frac{P(h)}{Q(h)}dh$, with probability at least $1-\sigma$ over $S_m$, for all $Q \in \mathcal{M}_1(\mathcal{H})$, we have: 
\begin{equation}
	E_{h\sim P}R_S(h)\leq \frac{1}{\beta}E_{h\sim P}R_{S_m}(h)+\frac{KL(P||Q)+ln\frac{1}{\sigma}}{|S_m|\cdot2\beta(1-\beta)} ,\label{s}
\end{equation} 
here, $\mathcal{M}_1(\mathcal{H})$ represents the probability space of $\mathcal{H}$,  we restrict the $P$,$Q$ within the Gibbs distribution, where the distribution of predictions of any input is predicted by classifiers $h$ randomly sampled from its distribution \cite{ref20}.  The data-dependent priors suggest that using the data can find certain $Q$ that tighten this bound \cite{ref24}. It is noted that pre-trained models are actually 'found' using large-scale data, which can also serve as  priors. Inspired by these, we reshape (\ref{common}) to base it on the distribution of $P \in \mathcal{M}_1(\mathcal{H})$, define a distribution $Q(D_0)\in\mathcal{M}_1(\mathcal{H})$ trained on $D_0$ and derive the FT-LP objective:
	  \begin{equation}
	  		\label{kl}
	  		\textbf{(FT-LP)}\quad	\mathop{\arg\min}_{P\in\mathcal{M}_1(\mathcal{H})} E_{h\sim P}(E_{\{x,y\} \sim S}L_{DG}(h(x),y)) + KL(P||Q(D_0)).
	 \end{equation}
	
	FT-LP incorporates the pre-trained model as $Q(D_0)$ into the optimization objective of DG, requiring $P$ to continuously consider its distance from $Q(D_0)$ during optimization. In the subsequent sections, we present propositions aimed at demonstrating the validity of (\ref{kl}) by establishing that the pre-trained prior can effectively narrow the bound. Then, considering the conditions where only $h_0$ is available, we derive a tractable upper bound to lead to our implementation objective.

     \subsection{Tightening DG generalization bounds with large-scaled pre-trained priors}
      Our first step is to transform (\ref{common}) into a form that targets the distribution of classifiers. 
      Generally, the objectives $L_{DG}(\cdot)$ can be seen as optimizing both the task loss $R_S(h)$ and a loss $dist(S, T,\mathcal{H})$ that measures the distance between $S$ and $T$ on $h\in \mathcal{H}$. In this case, $R_T(h)$ can be controlled by the following upper bound:
      \begin{equation}
      	R_T(h)\leq R_S(h) + dist(S,T,\mathcal{H}) + \lambda_p, \quad \forall h\in\mathcal{H}.\label{dist}
      \end{equation}
       Here, $\lambda_p$ is a constant independent of $h$. (\ref{dist}) is not new and has been previously discussed. For example, $dist(S,T,\mathcal{H})$ could represent metrics like total variation distance, H-divergence, or H$\triangle$H divergence \cite{ref21,ref47,ref56}, all of which adhere to the form in (\ref{dist}). These metrics hold for any model, but $T$ is inaccessible for DG \cite{ref8}. Therefore, many studies make certain assumptions aiming to reduce this generalization error using different $L_{DG}(\cdot)$ \cite{ref2,ref15,ref21}.  We will discuss this further in the \textbf{Appendix} to confirm their equivalence.
   
      
     The above inequality only considers the case of a single $h$. If we are interested in a distribution $P\in \mathcal{M}_1(\mathcal{H})$, we can take the expectation on both sides with $P$, and have the following formula:
        \begin{equation}
      E_{h\sim P}R_T(h)\leq E_{h\sim P}R_S(h) + dist(S,T,P) + \lambda_p. \label{ex}
        \end{equation} 
       We denote $E_{h \sim P}dist(S, T,\mathcal{H})$ as $dist(S, T, P)$. Based on (\ref{ex}), we can utilize PAC-Bayes bounds, which originally could not be used in DG problems due to domain shifts. By utilizing  (\ref{ex}) along with  (\ref{s}), we can derive the following results:
      
      \textbf{Theorem 1.}  $\forall \beta,\sigma \in (0,1)$, $n \in \mathbb{N}$, $P \in \mathcal{M}_1(\mathcal{H})$, $S_m \subset S$. With probability at least $1-\sigma$ over $S_m$, for all $Q \in \mathcal{M}_1(\mathcal{H})$, we have: 
      \begin{equation}
      E_{h\sim P}R_T(h)\leq \Phi(Q, P, S_m, T)= \frac{1}{\beta}E_{h\sim P}R_{S_m}(h)+\frac{KL(P||Q)+ln\frac{1}{\sigma}}{|S_m|\cdot2\beta(1-\beta)}+dist(S,T,P)+\lambda_p .\label{d}
      \end{equation}
      
  	While we employ a specific PAC-Bayes bound, (\ref{ex}) establishes connections between losses from different domains through distance definitions. This allows us to employ different PAC-Bayes bounds. Typically, we set $Q$ as a simple distribution, like the standard Gaussian, due to a lack of prior knowledge. However, it may not be the optimal choice here. Since Theorem 1 holds for any prior $Q$, the more information it provides, the better the performance of the posterior \cite{ref41,ref42}. To address this, consider a fixed learning rule for the posterior $P \Rightarrow P(S_m)$ \cite{ref23}, where $P(S_m)$ is the posterior learned from $S_m$. We aim for an "optimal" prior, denoted as $Q^*$, which represents the expected posterior over the source domain distribution $Q^* = E_{S_m\sim S}(P(S_m))$ \cite{ref41}. This $Q^*$ can effectively minimize our KL divergence term since $E_{S_m\sim S}KL(P(S_m)||Q^*) = 0$ \cite{ref24,ref41}. 
  	


      The aforementioned "optimal" distribution relies on an unknown $S$, making it impractical in real applications. Therefore, a simple strategy to leverage this method involves partitioning the training dataset \cite{ref23,ref48}. Let "$\backslash $" be the complement operation, and we randomly select a slice of the training data, denoted as $S_m^J$, where $J$ indicates the index of the chosen slice. We first train a data-dependent prior $Q(S_m^J)$, and then train our posterior distribution using $S_m \backslash S_m^J$,  we  replace $E_{S_m \sim S}$ in $Q^*$ with $E_JE_{S_m^J\sim S_m}$ to create an applicable prior, which can be written as: 
      
      \textbf{Theorem 2.}  $\forall \beta,\sigma \in (0,1)$, $n \in \mathbb{N}$, $P \in \mathcal{M}_1(\mathcal{H})$, $S_m \subset S$ and random index $J$ sampled from  $|S_m|$.  With probability at least $1-\sigma$ over $S_m$, for all $S_m^J$, we have: 
      \begin{equation}
      	E_{h\sim P}R_T(h)\leq E_J(
      \frac{1}{\beta}E_{h\sim P}R_P(S_m\backslash S_m^J)+\frac{KL(P||Q(S_m^J))+\log \frac{1}{\sigma}}{|S_m\backslash S_m^J|2\cdot\beta(1-\beta)}+dist(S,T,P)+\lambda_p) . \label{dependent}
      \end{equation}

         Under certain conditions, this approach can yield a tighter PAC-Bayes bound on expectation over $J$ \cite{ref23,ref48}. While in DG, these methods might not be optimal. The goal of DG is the unknown $T$, so the above $Q^*$ may not be optimal. Additionally, training the prior requires algorithms the same as the posterior. However, DG algorithms  \cite{ref15,ref16, Mixup} theoretically demand enough data to improve model generalization ability, which can be impractical with $S_m^J$. Therefore, these DG algorithms can lead to priors containing incorrect information and misguided optimization. We need other methods to estimate the priors.

      
      Since DG methods nowadays rely on publicly available pre-trained models \cite{resnet, clip}. We employ a similar method to the data-dependent prior, using $E_{D_0}(Q(D_0))$ to approximate the ideal prior. Here, $Q(D_0)$ may have different learning rules to $P$. $D_0$ is independent of $S$ or $T$, given $E_{D_0}(Q(D_0)) \in \mathcal{M}_1(\mathcal{H})$, we can derive a new bound:
      
      
      \textbf{Theorem 3.}  $\forall \beta,\sigma \in (0,1)$, $n \in \mathbb{N}$, $P \in \mathcal{M}_1(\mathcal{H})$, $S_m \subset S$. With probability at least $1-\sigma$ over $S_m$, for all $Q(D_0) \in \mathcal{M}_1(\mathcal{H})$, we have: 
       \begin{equation}
       	E_{h\sim P}R_T(h)\leq \frac{1}{\beta}E_{h\sim P}R_{S_m}(h)+\frac{E_{D_0}KL(P||Q(D_0))+ln\frac{1}{\sigma}}{|S_m|\cdot2\beta(1-\beta)}+dist(S,T,P)+\lambda_p. \label{th2}
       \end{equation}
     We provide a proposition below indicating that utilizing the distribution learned from large-scale pre-training as a prior $Q(D_0)$ is superior to that using partial training data. First, we restrict our priors in a family $\mathcal{F} \in \mathcal{M}_1(\mathcal{H})$, where $Q(D_0),Q(S_m^J) \in \mathcal{F} $ and we define the information gain difference $D_{\mathcal{F}}(h,S_m|D_0,J)$ , the risk bias $B_r(h,S_m|J) $ and the distance bias $B_d(h,S_m|J)$ as follows:
      
      
      \begin{equation}
      D_{\mathcal{F}}(h,S_m|D_0,J)=E_{S_m \sim S}(\frac{E_{D_0}KL(P||(Q(D_0)))}{|S_m|}-\frac{KL(P||E_J(Q(S)|S_m^J))}{|S_m\backslash S_m^J|}).
       \end{equation}
       \begin{equation}
       	B_r(h,S_m|J) = E_{S_m \sim S}E_{h \sim P}(R_{S_m\backslash S_m^J}(h)-R_{S_m}(h)),
       \end{equation}
        \begin{equation}
        	B_d(h,S_m|J) = E_{S_m \sim S}(dist(S_m\backslash S_m^J,T,P)-dist(S_m,T,P)).
        \end{equation}
       Notice that here we are using $dist(S_m, T, P)$ to approximate $dist(S, T, P)$, which typically involves a new parameter $\lambda_p^{S_m}$ related to the data volume $|S_m|$ \cite{DANN,ref58}. However,  if we fix the sampling quantity $|S_m^J|$, this will not affect the discussion. We provide the following proposition:
       
      \textbf{Proposition 1.}  $\forall \beta,\sigma \in (0,1)$, $n \in \mathbb{N}$, $P \in \mathcal{M}_1(\mathcal{H})$, $S_m \in S$. Suppose $|S_m| = m $, where all $J$ are randomly  and $|S_m\backslash S_m^J| = n < m$. Condition on $J$ and $S$, then $E_JE_{S_m \sim S}\Phi(Q(S),P(S_m^J\backslash S),S,T) \geq E_{D_0}E_{S_m \sim S}\Phi(Q(D_0),P(S_m^J),S,T)$ if and only if :
       \begin{equation}
       	\begin{split}
       	D_{\mathcal{F}}(h,S_m|D_0,J) \leq 2(1-\beta) B_r(h,S_m|J)+\beta(1-\beta)  	B_d(h,S_m|J) +\frac{ln\frac{1}{\sigma}}{n}\frac{n}{m-n} .
       	\end{split}
        \end{equation}
        The conditions mentioned above seem complex, requiring us to ensure that the information obtained from the $Q(D_0)$ is greater than that obtained from $Q(S_m^J)$. Additionally, these losses are not easily computed directly. Hence, we present a scenario that satisfies this proposition. We confine both $P$ and $Q$ to follow Gaussian distributions. Let $Q'$ denote the ideal model distribution capable of accurately predicting all data from $S$ and $T$. By employing $Q'$ as the prior, we can obtain the tightest bound of Theorem 1. Specifically, when $P=Q'$, all terms in Theorem 1 equate to zero, thereby attaining the minimum value. Let $Q'= N(\mu',\Sigma')$ and $Q=N(\mu,\Sigma)$. The following proposition explores the deviation from the ideal bound when substituting $Q'$ with an arbitrary $Q$:
        
        \textbf{Proposition 2.}  $\forall \beta,\sigma \in (0,1)$, $n \in \mathbb{N}$, $P,Q',Q \in \mathcal{M}_1(\mathcal{H})$, $P,Q',Q$ follow Gaussian distributions, $S_m \subset S$, assume $KL(P||Q)$ is differentiable with respect to $Q$ at $Q'$. When we using $KL(P||Q)$ to approximate the $KL(P||Q')$, with probability at least $1-\sigma$ over $S_m$, we have: 
        \begin{equation}
        	\begin{split}
        		E_{h\sim P}R_T(h)&\leq \frac{1}{\beta}E_{h\sim P}R_{S_m}(h)+\frac{KL(P||Q')+ln\frac{1}{\sigma}}{|S_m|\cdot2\beta(1-\beta)}+dist(S,T,P)+\lambda_p \\&
        		-C_1'||\mu-\mu^{'}||^2-C_2'tr(\Sigma-\Sigma^{'})^2 \label{d3}
        	\end{split}
        \end{equation}
        Here,  $tr(\cdot)$ represents getting the trace of the matrix, $C_1',C_2'$ are related to the $P$ and are bounded due to the properties of the Gaussian distribution. From (\ref{d3}), it is clear that when approximating with $KL(P||Q)$, the deviation from the ideal boundary mainly depends on the difference between $Q$ and $Q'$. Since controlling $C_1'$ and $C_2'$ is challenging, we only need to ensure that $KL(Q||Q')$ is minimized to approach the ideal bound.  
        
        Therefore, we can assume that $E_{D_0}KL(Q'||Q(D_0))\leq E_{J}KL(Q'||Q(S_m^J))$ to ensure Proposition 1 holds. While computing $Q'$ is challenging, We believe that large-scale pre-trained priors $Q(D_0)$ have encountered many distributions, possibly similar to that of $S$, which allows them to be closer to $Q'$. In contrast,  data-dependent priors $Q(S_m^J)$ rely on known $S_m$, but even a $Q$ that is entirely correct on $S$ might not necessarily be close to $Q'$. Hence, we believe this assumption is valid for large-scale pre-trained models.

     \subsection{Implementation for FT-LP }

     We have shown that under specific conditions, incorporating large-scale pre-trained models as priors aids in enhancing the generalization capability of the posterior. A simple approach is to optimize FT-LP with a Bayesian model $P$:
     \begin{equation}
     \mathop{\arg\min}_{P\sim \mathcal{M}_1(\mathcal{H})} E_{h\sim P}L_{DG}(h(x),y) + \gamma  KL(P||Q(D_0)). \label{ff}
     \end{equation}
      The parameter  $\gamma$ are adjustable hyperparameter derived from $\beta$ and $\sigma$ in (\ref{d}). However, achieving (\ref{ff}) poses significant challenges for DG. Firstly, optimizing Bayesian models such as ResNet50 \cite{resnet} for DG tasks is challenging and time-consuming due to their architectures \cite{ref25,ref49}. Secondly, finding suitable Bayesian priors is challenging because publicly available models often contain only weights without distributions \cite{clip,resnet}.
     
     In light of this, we consider utilizing the Maximum A Posteriori Estimation to optimize FT-LP. Specifically, we search for parameter estimation that minimizes the (\ref{ff}). To simplify the computation, we restrict both the prior and posterior distributions to the Gaussian distribution family like Proposition 2, i.e., $Q = N(\mu(h_0),\Sigma_{h_0}),P = N(\mu(h),\Sigma_{h})$. We reformulate the optimization objective as follows:
     \begin{equation}
     	\begin{split}
      &\mathop{\arg\min}_{h\sim \mathcal{H}}E_{(x,y)\sim S_m}L_{DG}(h(x),y)+\gamma (||\mu(h)-\mu(h_0)||^2_{\Sigma_{h_0)}^{-1}}+tr(\Sigma^{-1}_{h_0}\Sigma_{h})).  \label{r1}
       	\end{split}
      \end{equation}
     Through (\ref{r1}), we observe that computing the distance between two weights involves the covariance $\Sigma_{h_0}$ of the pre-trained distribution, which is typically challenging to obtain. One approach is to employ methods during pre-training to acquire the covariance from upstream data\cite{ref45,ref50}. However, for downstream tasks, this would introduce a considerable computational burden, especially considering that DG often assumes limited data available\cite{domainbed}. Therefore, we simply set the mean $\mu$ as the model weights: $\mu(h) = h, \mu(h_0) = h_0$, and use linear encoders $G$ encoding $h,h_0$ to get the covariance, with the optimization objective being the constraint terms in (\ref{r1}). So $Q(D_0) = N(h_0,G(h_0))$ and  $P = N(h,G(h))$.  Given a training model $h$ with only its pre-trained weights $h_0$ available, our implementation for FT-LP is:
        \begin{equation}
    	\begin{split}
    		&\mathop{\arg\min}_{h\sim \mathcal{H}}E_{(x,y)\sim S_m}L_{DG}(h(x),y)+\gamma (||h-h_0||^2_{G(h_0)^{-1}}+tr(G(h_0)^{-1}G(h))),\\&\qquad \qquad s.t.\ G \in \mathop{\arg\min}_{G}||h-h_0||^2_{G(h_0)^{-1}}+tr(G(h_0)^{-1}G(h)) \label{r2}
    	\end{split}
    \end{equation}
    We argue that this simplification does not compromise the assumption of independence between prior and source domains. The  objective of $G(\cdot)$ is to minimize the KL term while keeping prior weights fixed, regardless of the model performance. Notably, as the last layer of the model is typically task-specific, we employ a standard Gaussian distribution prior to our last layer. Details of our algorithm can be found in Algorithm \ref{alg:AOA}.
    
    \begin{algorithm}[h]
    	\caption{ Implementation for Fine-tuning with large-scale pre-trained Priors}
    	\label{alg:AOA}
    	\renewcommand{\algorithmicrequire}{\textbf{Input:}}
    	\renewcommand{\algorithmicensure}{\textbf{Output:}}
    	\begin{algorithmic}[1]
    		\REQUIRE Pre-trained model $h_0$, classifier $h$, training data $S_m$, update count $c$, training time $C$, scale parameter $\gamma$, batch size $B$, sample times $T$, covariance encoder $G$, arbitrary DG loss function $L_{DG}$. 
    		\ENSURE learned classfier $h_C$    
    		
    		\STATE Initalize $h$ with $h_0$ , random initalize $G$
    		\WHILE{$c\leq C$}
    		\STATE From $S_m$ randomly select $(X,Y)=\{x_i,y_i\}_{i=1}^{B}$ 
    		\STATE $L = \sum_{i=1}^{B}(L_{DG}(h_c(x_i),y_i) +\gamma (||h-h_0||^2_{G(h_0)^{-1}}+tr(G(h_0)^{-1}G(h))) $
    		\STATE update $h_c,\Sigma$ to minimize $L$
    		\STATE $c = c +1$ 
    		\ENDWHILE
    	\end{algorithmic}
    \end{algorithm}

\section{Experienments}
\subsection{Experiment setups and implementation details}
\textbf{Experiment setups and datasets.} We utilize the Domainbed evaluation protocols \cite{domainbed} for a fair comparison across five benchmarks:  \texttt{PACS} \cite{PACS} (4 domains, 7 classes, and 9,991 images), \texttt{VLCS}\cite{VLCS} (4 domains, 5 classes, and 10,729 images), \texttt{Office} \cite{Office} (4 domains, 65 classes, and 15,588 images), \texttt{TerraIncognita} \cite{Terra} (4 domains, 10 classes, and 24,788 images), and \texttt{DomainNet} \cite{Domain} (6 domains, 345 classes, and 586,575 images). The model selection, hyperparameter (HP) search, and data splits follow the Domainbed framework, utilizing validation within the source domain. Notice here we set a simple HP search space since it requires heavy computation resources.

\textbf{Evaluation protocol.}  All methods are validated through leave-one-out cross-validation. Each experiment designates one domain as the target, with the others as source domains. The final results average across three different training-validation splits, and the reported accuracy is the mean of all possible out-of-domain settings.

\textbf{Implementation details.} We selected 11 algorithms aiming to minimize $dist(S, T, \mathcal{H})$ from various perspectives. A pre-trained ResNet-50 \cite{resnet} from ImageNet \cite{ImageNet} serves as the prior and the backbone for all models. We set $R$ as cross-entropy for every model, and are trained using Adam optimizer with search space of [8e-5, 5e-5, 3e-5, 1e-5]. For HP unrelated to the algorithms, such as dropout rate and batch size, we refer to \cite{ref31} for search. For algorithm-specific hyperparameters, those already reported are directly used, while others are searched within default parameters provided by Domainbed, expanded by a factor $\lambda \in (100, 10, 1, 0.1, 0.01)$. To reduce computational load and demonstrate the stability of our method, these hyperparameters will be searched and fixed without incorporating FT-LP. Subsequently, we combine the FT-LP with these algorithms.  The search space for $\gamma$ is $\{[10^{i},0.5*10^{i}]|i=-8,-7,...,2\}$ since we fix other hyperparameters. We present results with our FT-LP algorithm and results without it. 

\subsection{Main results}

\begin{table}[]
	\centering
	\setlength{\tabcolsep}{3.3pt}
	\renewcommand{\arraystretch}{1.08}
	\caption{The main result, the top half displays results without FT-LP, while the bottom half shows the combined ones. Here, {\color{red}$\uparrow$} denotes an increase in accuracy after combining with FT-LP. Conversely, {\color{blue}$\downarrow$} indicates a decrease. The improvement in the table means the average accuracy improvement across five datasets compared to the original.}
	\label{tb1}
	\begin{tabular}{lrlrlrlrlrl|c}
		\hline
		\multicolumn{1}{c}{\textbf{Algorithm}} & \multicolumn{2}{l}{\texttt{PACS}}               & \multicolumn{2}{l}{\texttt{VLCS}}               & \multicolumn{2}{l}{\texttt{Office}}         & \multicolumn{2}{l}{\texttt{TerraInc}}         & \multicolumn{2}{l|}{\texttt{DomainNet}}         & Avg.     \\ \hline
		\multicolumn{1}{l|}{$\mathrm{MMD}$ \cite{MMD}}               & 84.7          &                        & 77.5          &                        & 66.3          &                        & 44.2          &                      & 23.4          &                        & 58.8     \\
		\multicolumn{1}{l|}{$\mathrm{IRM}$ \cite{ref15}}               & 83.5          &                        & 78.5          &                        & 64.3          &                        & 47.6          &                      & 33.9          &                        & 61.6     \\
		\multicolumn{1}{l|}{$\mathrm{VREx}$ \cite{ref16}}              & 84.9          &                        & 78.3          &                        & 66.4          &                        & 46.4          &                      & 33.6          &                        & 61.9     \\
		\multicolumn{1}{l|}{$\mathrm{CDANN}$ \cite{CDANN}}             & 82.6          &                        & 77.5          &                        & 65.8          &                        & 45.8          &                      & 38.3          &                        & 62.0       \\
		\multicolumn{1}{l|}{$\mathrm{DANN}$ \cite{DANN}}              & 83.6          &                        & 78.6          &                        & 65.9          &                        & 46.7          &                      & 38.3          &                        & 62.6     \\
		\multicolumn{1}{l|}{$\mathrm{Mixup}$ \cite{Mixup}}             & 84.6          &                        & 77.4          &                        & 68.1          &                        & 47.9          &                      & 39.2          &                        & 63.4     \\
		\multicolumn{1}{l|}{$\mathrm{ERM}$ \cite{domainbed}}               & 85.5          &                        & 77.3          &                        & 66.5          &                        & 46.1          &                      & 43.8          &                        & 63.9     \\
		\multicolumn{1}{l|}{$\mathrm{SagNet}$ \cite{SagNet}}            & 85.9         &                        & 77.8          &                        & 68.1          &                        & 48.6         &                      & 40.3          &                        & 64.2     \\
		\multicolumn{1}{l|}{$\mathrm{RIDG}$ \cite{RIDG}}            & 84.7          &                        & 77.8          &                        & 68.6          &                        & 47.8          &                      & 41.9          &                        & 64.2     \\
		\multicolumn{1}{l|}{$\mathrm{CORAL}$ \cite{CORAL}}             & 86.2          &                        & 78.8          &                        & 68.7          &                        & 47.6          &                      & 41.5          &                        & 64.5     \\
		\multicolumn{1}{l|}{$\mathrm{SAGM}$ \cite{SAGM}}              & 86.4          &                        & 79.1          &                        & \textbf{69.4}          &                        & 48.0          &                      & 44.8          &                        & 65.6     \\ \hline
		\multicolumn{11}{l|}{\textit{Combined with} FT-LP}                                                                                                                                                                                                       & Improve. \\ \hline
		\multicolumn{1}{l|}{$\mathrm{MMD}$}               & 84.7          &                        & 78.4          &\small\color{red} $\uparrow 0.9$   & 68.4          &\small\color{red} $\uparrow 2.1$   & 47.5          &\small\color{red} $\uparrow 3.3$ & 43.1          &\small\color{red} $\uparrow 19.7$  & $\uparrow$ 5.6      \\
		\multicolumn{1}{l|}{$\mathrm{IRM}$}               & 85.4          &\small\color{red} $\uparrow 1.9$   & 78.2          &\small  \color{blue} $\downarrow 0.3$ & 67.1          &\small\color{red} $\uparrow 2.8$   & 48.6          &\small\color{red} $\uparrow 1.0$ & 41.9          &\small\color{red} $\uparrow 8.0$   & $\uparrow$ 2.7      \\
		\multicolumn{1}{l|}{$\mathrm{VREx}$}              & 86.4          &\small\color{red} $\uparrow 1.5$   & 78.5          &\small\color{red} $\uparrow 0.2$   & 67.2          &\small\color{red} $\uparrow 0.8$   & 47.3          &\small\color{red} $\uparrow 0.9$ & 41.8          &\small\color{red} $\uparrow 8.2$   & $\uparrow$ 2.3      \\
		\multicolumn{1}{l|}{$\mathrm{CDANN}$}             & 84.0          &\small\color{red} $\uparrow 1.4$   & 77.9          &\small\color{red} $\uparrow 0.4$   & 67.8          &\small\color{red} $\uparrow 2.0$   & 48.2          &\small\color{red} $\uparrow 2.4$ & 38.5          &\small\color{red} $\uparrow 0.2$   & $\uparrow$ 1.3      \\
		\multicolumn{1}{l|}{$\mathrm{DANN}$}              & 84.4          &\small\color{red} $\uparrow 0.8$   & 79.0          &\small\color{red} $\uparrow 0.4$   & 67.8          &\small\color{red} $\uparrow 1.9$   & 49.0          &\small\color{red} $\uparrow 2.3$ & 38.6          &\small\color{red} $\uparrow 0.3$   & $\uparrow$ 1.2      \\
		\multicolumn{1}{l|}{$\mathrm{Mixup}$}             & 84.8          &\small\color{red} $\uparrow 0.2$   & 79.0          &\small\color{red} $\uparrow 1.6$   & 68.2          &\small\color{red} $\uparrow 0.1$   & 48.1          &\small\color{red} $\uparrow 0.2$ & 43.3          &\small\color{red} $\uparrow 4.1$   & $\uparrow$ 1.3      \\
		\multicolumn{1}{l|}{$\mathrm{ERM}$}               & 86.2          &\small\color{red} $\uparrow 0.7$   & 78.8          &\small\color{red} $\uparrow 1.5$   & 68.7          &\small\color{red} $\uparrow 2.2$   & 48.2          &\small\color{red} $\uparrow 2.1$ & 43.4          &\small  \color{blue} $\downarrow 0.4$ & $\uparrow$ 1.2      \\

		\multicolumn{1}{l|}{$\mathrm{SagNet}$}            & 85.8          &\small  \color{blue} $\downarrow 0.1$ & 78.8          &\small\color{red} $\uparrow 1.0$   & 68.1          &                        & \textbf{50.0} &\small\color{red} $\uparrow 1.4$ & 43.4          &\small\color{red} $\uparrow 3.1$   & $\uparrow$ 1.0      \\
		
		\multicolumn{1}{l|}{$\mathrm{RIDG}$}            & 85.5          &\small  \color{red} $\uparrow 0.8$ & 78.7          &\small\color{red} $\uparrow 0.9$   &  68.5  &\small  \color{blue} $\downarrow 0.1$                                & 48.0 &\small\color{red} $\uparrow 0.2$ & 42.0          &\small\color{red} $\uparrow 0.1$   & $\uparrow$ 0.4      \\

		\multicolumn{1}{l|}{$\mathrm{CORAL}$}             & 86.3          &\small\color{red} $\uparrow 0.1$   & 79.0          &\small\color{red} $\uparrow 0.2$   & 68.5          &\small  \color{blue} $\downarrow 0.2$ & 48.5          &\small\color{red} $\uparrow 0.9$ & 44.3          &\small\color{red} $\uparrow 2.8$   & $\uparrow$ 0.8      \\
		\multicolumn{1}{l|}{$\mathrm{SAGM}$}              & \textbf{86.7} &\small\color{red} $\uparrow 0.3 $  & \textbf{79.2} &\small\color{red} $\uparrow 0.1$   & \textbf{69.4} &                        & 48.6          &\small\color{red} $\uparrow 0.6$ & \textbf{44.9} &\small\color{red} $\uparrow 0.1$   & $\uparrow 0.2 $      \\ \hline
	\end{tabular}
\end{table}
Table \ref{tb1} displays the accuracy of the models on Domainbed. To highlight the contrast between combined FT-LP and uncombined results, we excluded variance terms from the table, with corresponding details provided in the \textbf{Appendix}. In the table, an upward arrow signifies an accuracy improvement, while a downward arrow indicates a decrease. Bold entries denote the best results per dataset. Note that SAGM is a very effective method \cite{SAGM}, and we can still make improvements to it. We can also see most algorithms demonstrate enhanced accuracy when combined with FT-LP, with notable improvements observed. Even in cases of slight performance decline, the decrease is minimal  (maximum decrease is 0.4). Across all datasets, there is a consistent enhancement, underscoring the stability and effectiveness of our approach.

Our penalty term notably enhances the performance of basic ERM on most datasets, yielding an average increase of 1.2 percentage points.  Despite neither ERM training nor our method targeting the reduction of the source-target domain difference, we effectively boost model accuracy on $T$ without optimizing $dist(S,T,\mathcal{H})$.   ERM has proven to be highly competitive, with many DG methods struggling to surpass it. Incorporating our approach leads to significant improvements in generalization ability, surpassing the baseline results. This underscores the superiority of our fine-tuning method, especially for TerraIncognita and DomainNet datasets. In DomainNet, methods such as MMD increased from 23.4\% to 43.1\%, IRM from 33.8\% to 41.9\%, and VREx from 33.6\% to 41.8\%, while in TerraIncognita, all methods combined with FT-LP exhibit noticeable enhancements.  In total, every method combined with FT-LP has been improved, and the improvement is obvious, from 0.2\% to 3.3\%. These results demonstrate the effectiveness of our method on this dataset, with most models benefiting from the weights of pre-trained models.

\subsection{Visualization experiment}
   \begin{figure}
	\centering
	\includegraphics[width=1\textwidth]{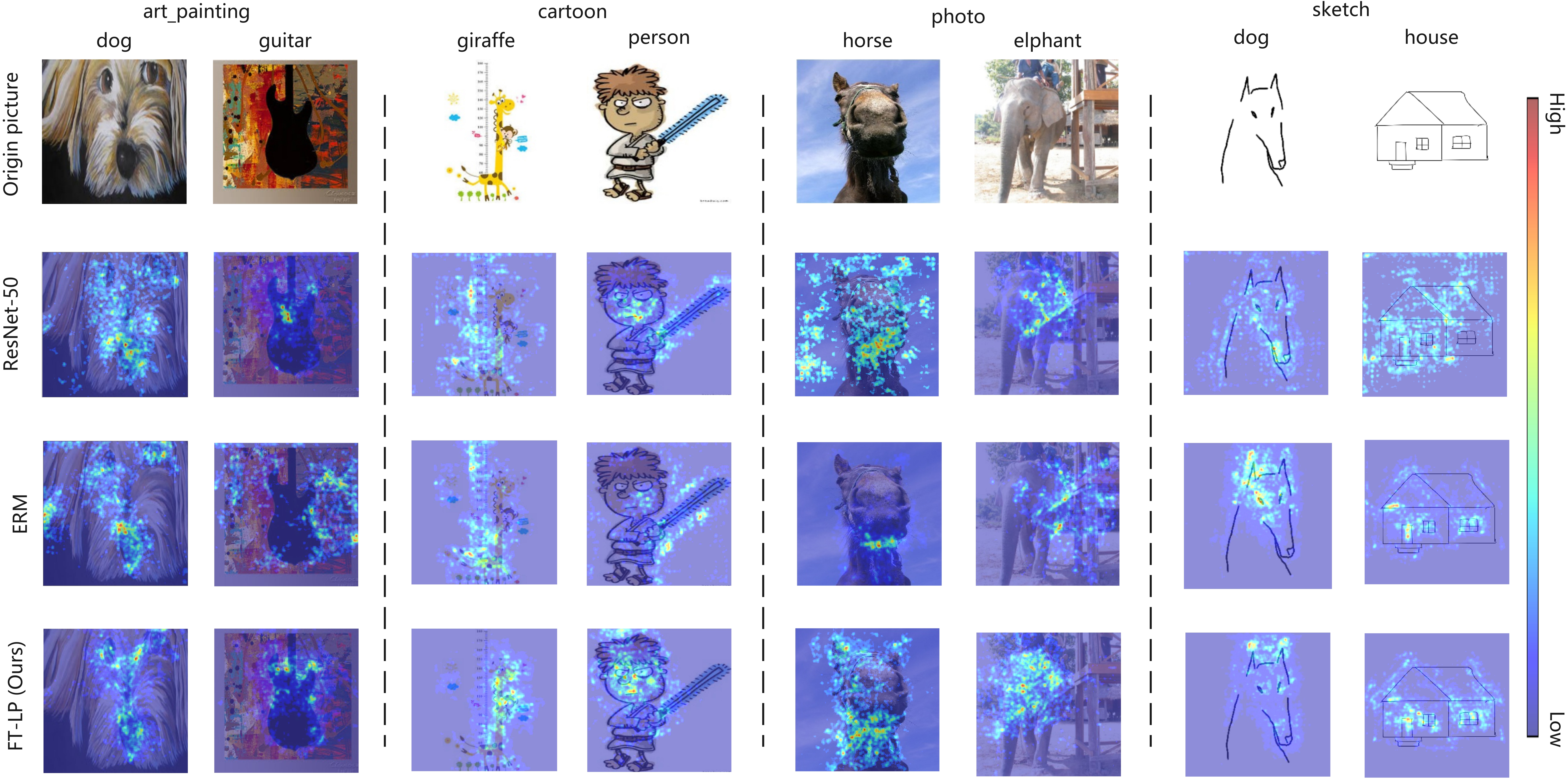}
	\caption{In the image, each pair of columns displays visualization results for different testing domains. Each column represents a specific model. Redder areas in the image indicate where the model focuses more on classification.}
	\label{fig3}
\end{figure}

We employ visual experiments to validate our key idea: pre-trained models, trained on diverse distributions, offer valuable knowledge to mitigate domain shifts, which are challenging for domain generalization methods to learn alone. Through visualization of the PACS dataset, we showcase how our approach leverages pre-trained model knowledge. We compare three models: a pre-trained ResNet-50 model, serving as initialization; a model trained only with ERM; and an ERM model combined with our FT-LP objective.  These models correspond to the second to fourth rows in Fig.(\ref{fig3}). The ERM and the FT-LP models correspond to the 8th and 19th rows in table \ref{tb1}, respectively.

In Fig.(\ref{fig3}), we illustrate the attention of models towards images, where a higher degree of redness signifies a greater level of attention. The purely pre-trained model focuses on various features due to its lack of exposure to the source domain. For instance, in the first column, it attends to both texture features like fur and features such as eyes and nose, thus it focuses on both invariant and non-invariant features. Conversely, the ERM, without any feature constraints, tends to capture more texture or background features, overlooking crucial features such as the outline of the object. For example, in the second column, although the guitar is primarily black, the ERM model fails to recognize it and instead focuses on the colorful textures nearby.

FT-LP, influenced by both pre-trained model weights and ERM, Its focus area overlaps with both, but the difference lies in being constrained by pre-trained weights. This means it considers a wider range of features during training, rather than immediately narrowing down to specific ones. As a result, it can ultimately select features more conducive to generalization. Since this constraint is domain-agnostic, we can conclude that this enhancement can be applied to other models, as confirmed by the main results.



\section{Discussion and Limitation}
\label{dissc}
\textbf{Non-Bayesian optimization strategies} Based on our previous discussion, it is worth noting that while our approach is rooted in the PAC-Bayes theory, we opted for MAP estimation instead of traditional Bayesian optimization methods due to various constraints. The latter typically requires a well-defined prior distribution and longer optimization times. It is important to acknowledge that Bayesian optimization has advantages, such as improving the accuracy similar to ensemble methods. However, this may compromise the ease of combinability inherent in FT-LP. Obtaining a well-trained Bayesian model is challenging, and the stochasticity introduced by the model distribution may render certain domain generalization methods incompatible, such as IRM and VREx, which primarily adjust classifier weights to obtain invariant features.  

\textbf{FT-LP only use one pre-trained model.} In our paper, $Q$ is only regarded as a single pre-trained model for the training model $P$. While in reality, we may have multiple pre-trained models $ Q(D_0), Q(D_1),...,Q(D_n)$. We should be able to utilize multiple priors simultaneously to further enhance the generalization capability in this case. However, the mentioned approach is likely to be complex and involve more parameters. Given the complexity of neural networks, simple linear combinations like the mixture of Gaussian distributions: $Q = w_1 Q(D_1) + w_2 Q(D_2) + ... + w_n Q(D_n)$ may lead to a loss of information in priors, as these parameters do not originate from the same optimization process.

\textbf{FT-LP requires additional memory and computational resources.} FT-LP demands more memory and computational resources as it requires updating the encoder to acquire variance, thus increasing memory needs. Our encoders match the parameter count of the original model, effectively doubling it. However, it is worth noting that these extra parameters do not entail a complex optimization process, as the covariance between layers is independent. Therefore, we recommend utilizing at least a 24GB GPU for optimization training. Additionally, for models with a higher parameter count, employing more GPUs in parallel training is advisable.

\section{Conclusion}
In this paper, we argue that the existing DG methods default to employing pre-trained weight without considering its rich information. Therefore, we propose FT-LP to leverage these weights theoretically in fine-tuning and introducing a new generalization upper bound for DG. In theory, We verify that incorporating pre-trained models as priors within this bound effectively reduces the prediction error on the target domain.  In the implementation, when we only have pre-trained model weights, we loosen optimization constraints to obtain MAP solutions. Then, we employ encoders to encode these weights, stimulating the model distribution for the application of FT-LP. Consequently, FT-LP can combine with other models adhering to our assumptions to further bolster effectiveness. We conduct experiments across multiple models and datasets within Domainbed to evaluate FT-LP performance. 


\clearpage
\bibliographystyle{IEEEtran}
\bibliography{Reference}
\clearpage
\begin{appendices}

\section{Some Discussion on Theoretical Details}
\subsection{The validity of $dist(S,T,\mathcal{H})$}

In the manuscript, we mention Domain Generalization (DG) methods often aimed at reducing $dist(S,T,\mathcal{H})$ and $R_S(h)$. The theoretical foundation of these methods can be summarized by (\ref{dist}), encompassing inequalities involving the source domain $S$, target domain $T$, and hypothesis space $\mathcal{H}$.  For example, the objective of some methods is to decrease the distance in feature space between different source domains to align them with the target domain \cite{DANN,CDANN,MMD,CORAL}, while also trying to reduce the $R_S(h)$. They have theoretical upper bound like (\ref{dist}) which includes the $S,T,\mathcal{H}$. However, not all methods have such theoretical bounds. Therefore, we provide a more detailed discussion here regarding their equivalence. We restrict our discussion to binary classification problems, considering the existence of a function $f$ capable of correctly classifying samples in a specific domain: $f_S : X \rightarrow Y=[0,1]$, for all $(X,Y) \in S$. Here, $R_S(h)$ represents the task loss, defined as $E_{x \sim S}|h(x)-f_S(x)|$. In practical settings, losses like cross-entropy are often employed to achieve this objective.

Firstly, it is important to note that $dist(S,T,\mathcal{H})$ is a generalized term, encapsulating numerous possibilities. Two potential scenarios are as follows\cite{ref21}:

\textbf{Variation Divergence.} For any $\mathcal{H}$,the inequality holds:
\begin{equation}
	R_{T}(h)\leq R_{S}(h)+ + d_1(S,T)+min (E_{S}(|h^*_S(x)-h^*_T(x)|),E_{T}(|h^*_S(x)-h^*_T(x)|)).
\end{equation}

Here, $h^*_S(x)$ and $h^*_T(x)$ respectively represent the models that minimize the loss $R$ on $S$ and $T$. $d_1(S,T) = 2\sup_{X\in\mathcal{X}}(P_S(X)-P_T(X)))$.

\textbf{$\mathcal{H}\Delta \mathcal{H}$ divergence.} We set $\mathcal{H}\Delta \mathcal{H}$ hypothesis space as :
$g(x)\in \mathcal{H}\Delta \mathcal{H} \Rightarrow g(x)=h(x)\oplus h'(x)$, for some $h,h'\in \mathcal{H}$ ,$\oplus$ means the XOR operation.
And we can set the $d_{\mathcal{H}\Delta \mathcal{H}}(S,T)$ as 
\begin{equation}
d_{\mathcal{H}\Delta \mathcal{H}}(S,T)=2\sup_{h,h'\in \mathcal{H}}|P_{x\sim S}(h(x)\neq h'(x))-P_{x\sim T}(h(x)\neq h'(x))|
\end{equation}
then the inequality holds: 
\begin{equation}
			R_T(h)\leq R_S(h)+\frac{1}{2}d_{\mathcal{H}\Delta \mathcal{H}}(S,T)+\lambda. \label{dh} 
\end{equation}

Here, $\lambda$ is related to $S$, $T$, and $\mathcal{H}$, but independent of the specific $h$. $d_{\mathcal{H}\Delta \mathcal{H}}(S,T)$ represents the comparison between the current function $h$ and the function $h^*_{S,T}$ that minimizes the loss $R$ on both $S$ and $T$.

It can be observed that these inequalities establish the relationship between the loss on the target domain $T$ and the loss on the source domain $S$. However, in the DG setting, we cannot access arbitrary information from the target domain $T$, making it difficult to optimize through these inequalities. Therefore, DG methods make some assumptions about $h^*_{S,T}$, expecting it to exhibit certain characteristics so it can be found through source domains. One widely used assumption is that $h^*_{S,T}$ captures invariant features. The core of this assumption lies in defining an invariant feature $x_{inv}$, which determines the classification outcomes in the source and target domain. In other words, employing a model $h \in \mathcal{H}$ based on $x_{inv}$ allows:

\begin{equation}
	R_S(h(x_{inv}))=R_T(h(x_{inv}))\leq R_S(h(x)),\forall h \in \mathcal{H}
\end{equation}
When we plug this simple assumption into (\ref{dh}), it ensures that without utilizing $x_{inv}$ for prediction, the inequality $R_T(h)< R_S(h)+\frac{1}{2}d_{\mathcal{H}\Delta \mathcal{H}}(S,T)+\lambda$ holds. The equation holds only when $R_S(h(x_{inv}))=R_T(h(x_{inv}))$, indicating its validity when employing invariant features for prediction. Under this assumption, we notice that the model fulfills the inequality condition in (\ref{dh}) when it fails to identify invariant features. Conversely, it meets the equality condition only when invariant features are identified. In other words, finding invariant features is an indirect way to attempt to reduce the upper bound of (\ref{dist}).

To illustrate this point further, let us consider the Invariant Risk Minimization (IRM) assumption\cite{ref15}:

In an environment comprising multiple domains, we assume the existence of an invariant feature $x_{inv}$ that remains consistent across all domains and determines the values of the labels $Y$. In other words, $x_{inv}$ exhibits the same distribution across different domains.

This assumption can be formally expressed as: For all $i$ and $j$, it holds that $P(x_{inv} | S_i) = P(x_{inv} | S_j)$, where $S_i$ denotes the $i$-th domain. Building upon this, IRM posits that the search for invariant features is equivalent to the following definition:

\textbf{IRM Definition 3. \cite{ref15}} We say that a data representation $\Phi:\mathcal{X}\to\mathcal{Z}$  elicits an invariant predictor $w\circ\Phi$ across environments $\mathcal{E}$ if there is a classifier $w: \mathcal{Z} \to \mathcal{Y}$  simultaneously optimal for all environments, that is, $w\in \arg \min _{\bar{w} : \mathcal{Z} \to \mathcal{Y} }R^e( \bar{w} \circ \Phi )$  for all$ e\in \mathcal{E} $.

Here, we introduce a linear classifier $w$, which forms the classifier $w \circ \Phi = h$. IRM primarily assumes the existence of a linear classifier that achieves optimality across all domains. In other words, IRM imposes constraints on the optimal model $h^*_{S,T}$, asserting that it should adhere to IRM Definition 3. Under this assumption, we can verify that $\hat{h}$, satisfying $w\in \arg \min _{\bar{w} : \mathcal{Z} \to \mathcal{Y} }R^e( \bar{w} \circ \Phi )$, where $\hat{h} = w\circ \Phi$, maintains optimality across all domains, i.e., $\hat{h}=h^*_{S,T}$. Consequently, we obtain the conditions for equality in (\ref{dh}). In cases where these conditions are not fulfilled, a re-evaluation of (\ref{dh}) can show that only inequality is satisfied. This demonstrates the equivalence between the IRM objective and $dist(S,T,\mathcal{H})$.

Certainly, in addition to invariant features, there are many other assumptions about $h^*_{S,T}$ \cite{Mixup,SAGM}. However, DG methods argue that the classifier $\hat{h}$ found under their assumptions can fit the $R_S(\hat{h})=R_T(\hat{h})$, satisfying the equality conditions for (\ref{dist}). In cases where they are not equal, we can always use metrics like H-divergence to measure the difference between them. Therefore, we can generalize them as attempting to optimize both $dist(S,T,\mathcal{H})$ and $R_S(h)$ simultaneously. 

Overall, As models cannot access information from the target domain $T$ in domain generalization settings, researchers attempt to impose constraints on the ideal $h^*_{S,T}$, attempting to reduce $dist(S,T)$ equivalently by seeking such potentially optimal classifiers.
\subsection{An example for Proposition 1 and 2}
  \begin{figure}
	\centering
	\includegraphics[width=1\textwidth]{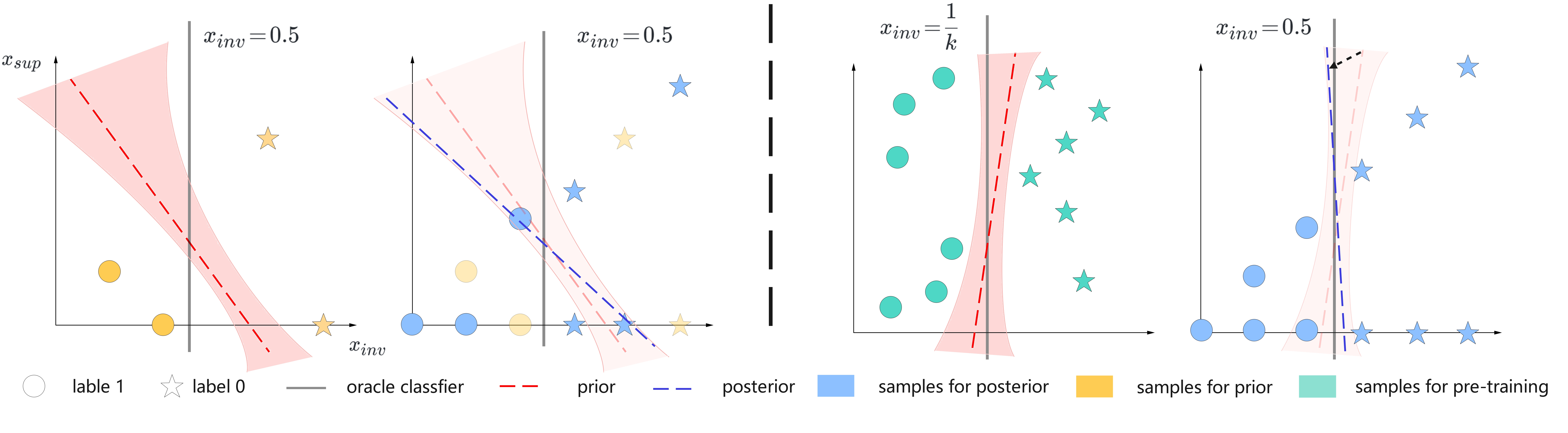}
	\caption{The horizontal and vertical axes represent $x_{inv}$ and $x_{sup}$ respectively,  different shapes represent different categories, and various colors indicate samples used for training different classifiers. The shaded regions depict the confidence intervals generated by prior covariance.  For clarity, we omit the confidence intervals of the posterior. The classifier that approaches vertical alignment is preferred.}
	\label{fig1}
\end{figure}

 Let us illustrate our propositions deeper with a simple linear classification example, which is adapted from \cite{ref51} to suit our context. Consider a scenario where our samples have two dimensions:
\begin{equation}
	\begin{split}
		X=(x_{inv}, &x_{sup}),  Y\leftarrow I(x_{inv}-\frac{1}{2}), P(x_{inv}=i)=\frac{1}{5}, i\in[0,0.2,0.4,0.6,0.8,1],
		\\&x_{sup}=W^e x_{inv},  W^e\in [0,1], W^e\sim \text{Bernoulli}(1-p^e). 
	\end{split}
\end{equation}
Here, $x_{inv}$ denotes invariant features for classification, while $x_{sup}$ represents spurious features.  Bernoulli$(a)$ indicates the probability of taking the value 1 as $a$ and the value 0 as $1-a$. Different training domains have varying values of $p^e$. We train Bayesian linear classifiers \(y=I(W_{inv}x_{inv}+W_{sup}x_{sup}+b)\), \(W_{sup}, W_{inv}, b\sim N(\mu,\sigma^2)\), using the same method for both prior and posterior training. The first two pictures of Fig.(\ref{fig1}) show one possible scenario. In this situation, many classifiers may achieve optimal results \cite {ref51}, indicating insufficient prior information.  We train the posterior with this data-dependent prior by minimizing the right part of (\ref{dependent}), the second figure shows one possible result. Constrained by the prior, the posterior likely remains within the shaded region, leading to inaccurate classification of training samples, making it challenging to achieve good generalization.

Then considering a large-scale dataset loosely associated with the desired task, we explore another classification scenario involving:
\begin{equation}
	\begin{split}
		&Y \leftarrow I(x_{inv}-\frac{1}{k}),  x_{inv}\sim N(\frac{1}{k},1),|k|>1,k\neq2
		\\&x_{sup}=W^e\cdot x_{inv}+\frac{e}{n}, W^e\in[0,1], W^e\sim \text{Bernoulli}(1-p^e),
	\end{split}
\end{equation}
\(n\) represents the environmental numbers, assuming it is sufficiently large to show the diversity of \(x_{sup}\) here.  We train classifiers the same as above. In third picture in Fig.(\ref{fig1}), with large samples of \(x_{sup}\), the prior may favor larger  \(W_{inv}\) and \(b\) for classification with small $\sigma$, driving \(W_{sup}\) towards 0. Considering the relevance of the problem, we use $W_{sup},W_{inv}$ to initialize the posterior, akin to using large-scale pre-trained models for initialization. The fourth picture shows one possible result where the posterior is shaped by the large-scale pre-training distribution, favoring a model with \(||\mu_{W_{inv}}||>>||\mu_{W_{sup}}||\), thereby enhancing its generalization ability. We can see that even across different tasks, knowledge from large-scale pre-trained models still aids in obtaining posterior with enhanced generalization.

\subsection{The effect of different $D_0$}

We also need to consider how varying distributions of pre-trained models affect outcomes. Relying solely on $KL(P||Q)$ is inadequate because the primary difference lies in the pre-training datasets here. Additionally, we are more concerned with the performance of the model on unknown $T$. A  $D_0$ that aligns more closely with the $T$ yields a better model. We approach the issue from the perspective of the pre-training dataset used.

\textbf{Proposition 3.} We define the Wasserstein distance as $w(P, Q)$ and denote the joint distribution of two distributions as $\gamma(P, Q)$. Given the distributions of two pre-training datasets $D_1$ and $D_2$, with an assumption of a non-empty intersection $D_{12}$ between them, and considering any arbitrary target domain distribution $T$, we can express:
\begin{equation}
	\left| w(D_1,T) - w(D_2,T) \right| \leq 2 \left( sup_{(x,y)\in \gamma(D_1,T)}|x-y| + sup_{(x,y)\in \gamma(D_2,T)}|x-y| \right).
\end{equation}
Proposition 2 highlights that the difference between two distances is mainly affected by the samples with the largest disparity in their distributions.  Keeping $D_2$ fixed and diversifying the distribution of $D_1$, we can see a  increase in $|w(D_1,T) - w(D_2,T)|$ . This suggests that $Q(D_1)$ is likely to have a more significant impact on $P$ than $Q(D_2)$, though it is unclear if the impact is positive or negative. For example, a pre-trained model, despite exposure to diverse image classification tasks, may not outperform others in medical image classification. If we ensure $w(D_1, T) \leq w(D_2, T)$ by selecting $D$ sufficiently close to $S$ when the distributions of $S$ and $T$ are close, then enhancing $D_2$ diversity becomes more beneficial, leading to further improvements in $P$ generalization capacity. Therefore, choosing the pre-trained model based on the characteristics of the existing source domain proves to be more advantageous.

\section{Proof}
\label{proof}
\subsection{Proof for Proposition 1}
First we calculate the $\Phi(Q(S),P(S_m^J\backslash S),S_m^J\backslash S,T)$ condition on $J$:
\begin{equation}
	\begin{split}
		&E_{S_m \sim S}\Phi(Q(S),P(S_m^J\backslash S),S,T)
		\\& = \frac{1}{\beta}E_{S_m\sim S}E_{h\sim P}R_P(S_m\backslash S_m^J)+\frac{E_{S_m\sim S}(KL(P||Q(S_m^J)))+\log \frac{1}{\sigma}}{|S_m\backslash S_m^J|2\cdot\beta(1-\beta)}\\&+E_{S}dist(S_m\backslash S_m^J,T,P)+\lambda_p
		\\& = \frac{1}{\beta}E_{S_m\sim S}E_{h\sim P}R_P(S_m\backslash S_m^J)+\frac{E_{S_m\sim S}(KL(P||E(Q(S_m)|S_J)))+\log \frac{1}{\sigma}}{|S_m\backslash S_m^J|2\cdot\beta(1-\beta)}\\&+E_{S}dist(S_m\backslash S_m^J,T,P)+\lambda_p , 
	\end{split}
\end{equation}

For $ E_{D_0}E_{S_m \sim S}\Phi(Q(D_0),P(S_m^J),S,T)$ it is the same:
\begin{equation}
	\begin{split}
		& E_{D_0}E_{S_m \sim S}\Phi(Q(D_0),P(S_m^J),S,T)
		\\& = \frac{1}{\beta}E_{D_0}E_{S_m\sim S}E_{h\sim P}R_P(S_m)+\frac{E_{D_0}E_{S_m\sim S}(KL(P||Q(D_0)))+\log \frac{1}{\sigma}}{|S_m|2\cdot\beta(1-\beta)}\\&+E_{S}E_{D_0}dist(S_m,T,P)+\lambda_p
		\\& = \frac{1}{\beta}E_{S_m\sim S}E_{h\sim P}R_P(S_m)+\frac{E_{S_m\sim S}E_{D_0}(KL(P||(Q(D_0))))+\log \frac{1}{\sigma}}{|S_m|2\cdot\beta(1-\beta)}+E_{S}dist(S,T,P)+\lambda_p, 
	\end{split}	
\end{equation}

then we suppose $J \neq \emptyset$, $E_{D_0}E_{S_m \sim S}\Phi(Q(D_0),P(S_m^J),S,T) \leq E_{S_m \sim S}\Phi(Q(S),P(S_m^J\backslash S),S,T)$ equals:
\begin{equation}
	\begin{split}
		&\frac{1}{\beta}E_{S_m\sim S}E_{h\sim P}R_P(S_m)+\frac{E_{S_m\sim S}E_{D_0}(KL(P||(Q(D_0))))+\log \frac{1}{\sigma}}{2m\cdot\beta(1-\beta)}+E_{S}dist(S,T,P)+\lambda_p \\&\leq  \frac{1}{\beta}E_{S_m\sim S}E_{h\sim P}R_P(S_m\backslash S_m^J)+\frac{E_{S_m\sim S}(KL(P||E(Q(S_m)|S_J)))+\log \frac{1}{\sigma}}{2n\cdot\beta(1-\beta)}\\&+E_{S}dist(S_m\backslash S_m^J,T,P)+\lambda_p,
			\end{split}	
	\end{equation}
	
it is equivalent to:
\begin{equation}
	\begin{split}
		&(1-\beta)E_{S_m\sim S}E_{h\sim P}R_P(S_m)+\frac{E_{S_m\sim S}E_{D_0}(KL(P||(Q(D_0))))+\log \frac{1}{\sigma}}{2m}\\&+\beta(1-\beta)E_{S}dist(S,T,P) \\&\leq  (1-\beta)E_{S_m\sim S}E_{h\sim P}R_P(S_m\backslash S_m^J)+\frac{E_{S_m\sim S}(KL(P||E(Q(S_m)|S_J))) +\log\frac{1}{\sigma}}{2n}\\&+\beta(1-\beta)E_{S}dist(S_m\backslash S_m^J,T,P),
		\end{split}	
\end{equation}

rearranging inequality:
\begin{equation}
	\begin{split}
		&E_{S_m\sim S}(\frac{E_{D_0}(KL(P||Q(D_0)))+\ln\frac{1}{\sigma} }{2m}-\frac{KL(P||E_J(Q(S|S_m^J)))+\ln\frac{1}{\sigma}}{2n}) \\& \leq (1-\beta)E_{S_m\sim S}E_{h\sim P}(R_{S_m\backslash S_m^J}(h)-R_{S_m}(h)) \\& + \beta(1-\beta)E_{S_m\sim S}(dist(S_m\backslash  S,T,P)-dist(S_m,T,P)), 
			\end{split}	
\end{equation}
then utilizing the previously mentioned abbreviations and take the expectation with respect to 
$J$ on both sides of the inequality get :
\begin{equation}
	\begin{split}
		D_{\mathcal{F}}(h,S_m|D_0,J) \leq 2(1-\beta) B_r(h,S_m|J)+\beta(1-\beta)  	B_d(h,S_m|J) +\frac{ln\frac{1}{\sigma}}{n}\frac{n}{m-n} . 
	\end{split}
\end{equation}

We have the results.
\subsection{Proof for  Proposition 2}

In the subsequent analysis, we constrain our distributions to Gaussian distributions. Let $f(\mu,\Sigma)=KL(P||Q)$ be a function of $Q$. We assume that this function is differentiable at $Q'$. Thus, we have the following derivatives:
\begin{equation}
	\begin{split}
		&\nabla_{\mu_{Q'}}f(\mu,\Sigma)=\Sigma^{-1}(\mu_P-\mu_{Q'})
		\\& \nabla_{\Sigma_{Q'}}f(\mu,\Sigma) = -\Sigma_{Q'}^{-1}(\mu_{Q'}-\mu_P)(\mu_{Q'}-\mu_P)^T\Sigma_{Q'}^{-1}-\Sigma_{Q'}^{-1}\Sigma_{P}\Sigma_{Q'}^{-1}-\Sigma_P^{-1}
	\end{split}
\end{equation}
For the KL divergence between two Gaussian distributions, utilizing the condition of differentiability at $Q'$, we can use a first-order Taylor expansion to obtain the following expression:
\begin{equation}
	\begin{split}
		KL(P||Q)=KL(P||Q')&+\nabla_{\mu}f(\mu,\Sigma)^T|_{\mu=\mu_c}(\mu-\mu_{Q'})\\&+tr(\nabla_{\Sigma}f(\mu,\Sigma)^T|_{\Sigma=\Sigma_C}(\Sigma-\Sigma_{Q'}))
	\end{split}
\end{equation}

At this point, our partial derivatives take on specific values at $\mu_c,\Sigma_C$. By observing the above expression, we can derive the following inequality:
\begin{equation}
	\begin{split}
		&KL(P||Q')+\nabla_{\mu}f(\mu,\Sigma)^T|_{\mu=\mu_c}(\mu-\mu_{Q'})+tr(\nabla_{\Sigma}f(\mu,\Sigma)^T|_{\Sigma=\Sigma_C}(\Sigma-\Sigma_{Q'}))
		\leq \\&KL(P||Q') + C_1||\mu-\mu_{Q'}||^2+C_2tr(\Sigma-\Sigma_{Q'})^2
	\end{split}
\end{equation}
We can make the above expression hold by choosing appropriate $C_1$ and $C_2$. We substitute this expression into Theorem 1 to get :

\begin{equation}
	\begin{split}
		E_{h\sim P}R_T(h)\leq& \frac{1}{\beta}E_{h\sim P}R_{S_m}(h)+\frac{KL(P||Q')+ln\frac{1}{\sigma}}{|S_m|\cdot2\beta(1-\beta)}+dist(S,T,P)+\lambda_p \\&
	 C_1'||\mu-\mu_{Q'}||^2-C_2'tr(\Sigma-\Sigma_{Q'})^2\label{d2}
	\end{split}
\end{equation}
\subsection{Proof for  Proposition 3}
For any $S,T$, suppose $D_1 \cap D_2 = D_{12} \neq \emptyset$, we have :
\begin{equation}
	\begin{split}
     |w(D_1,T)-w(S,T)| &\leq |w(D_1,D_{12})+w(D_{12},T)-w(S,D_{12})+w(D_{12},T)|
	\\& =|w(D_1,D_{12})+2w(D_{12},T)-w(S,D_{12})|
	 \\& \leq |w(D_1,D_{12})-w(S,D_{12})|+ 2sup_{(x,y)\in \gamma(D_1,T)}|x-y|.
	\end{split}
\end{equation}

Here, the first inequality is derived from the triangle inequality definition of the Wasserstein distance, applied within the non-empty set $D_{12}$. The second inequality is established based on the definition of the Wasserstein distance, asserting that the expected distance between two metrics is necessarily smaller than their maximum disparity. For $|w(D_2,T)-w(S,T)|$
, we have the same conclusion, so as a result:
\begin{equation}
	\left| w(D_1,T) - w(D_2,T) \right| \leq 2 \left( sup_{(x,y)\in \gamma(D_1,T)}|x-y| + sup_{(x,y)\in \gamma(D_2,T)}|x-y| \right).
\end{equation}

\section{Implementation details}

For simplicity, our covariance encoder $G$ is a basic linear layer, with input parameters of the pre-trained model and outputting covariance of the same size, utilized for computing the KL divergence term. Note here we restrict the $G(\cdot)$ as a diagonal matrix, assuming independence between the weights. While this may result in some loss of information, we still deem this simplification reasonable, as it significantly reduces the required computational resources and time.

In practical deployment, we first take the parameters $f_i$ with numerical values from each layer of the model and use the corresponding encoder to encode them, obtaining vectors of the same length as the covariance: $\Sigma_{f_i} = G(f_i)$. Each layer has its encoder. Note that a complex encoder could quickly escalate the space and computational resources required by the model, potentially rendering it impractical for deployment. More details can be found in the code.
%

\section{Details in Experiment}
\label{exp}
\subsection{Optimal parameter for FT-LP on each dataset}
As mentioned in the main text, our parameter search space includes: learning rate [8e-5, 5e-5, 3e-5, 1e-5], dropout[0.5,0.1,0], weight decay[1e-4, 1e-6, 0], $\gamma$ from 100 to 1e-8 divided by 5 for ERM, while others are $\gamma_2 = \lambda_2\gamma^*, \lambda_2\in [100,10,1,0.1,0.01]$ to reduce computation source. The learning rate for our encoder is fixed at 10 times the learning rate of the main model. We provide the optimal combinations of these parameters on the FT-LP under ERM in the table below for easy reproducibility. The remaining parameters not mentioned are set to the default values used in Domainbed\cite{domainbed}. Other model parameters will be released along with the code, which is currently being organized. All experiments were conducted on a single NVIDIA RTX 3090 GPU. Python: 3.8.17, PyTorch: 2.0.1, Torchvision: 0.15.2, CUDA: 11.7, CUDNN: 8500, NumPy: 1.22.0, PIL: 9.4.0.
\begin{table}[htbp]
	\centering
	\caption{The optimal parameter settings for ERM combined with FT-LP on each dataset.}
	\begin{tabular}{@{}lllll@{}}
		\toprule
		Dataset        & learning rate & dropout rate & weight decay & hyperparameter $\gamma$ \\ \midrule
		PACS           & 3e-5          & 0            & 0            & 5e-6                    \\
		VLCS           & 1e-5          & 0.5          & 1e-6         & 5e-3                    \\
		OfficeHome     & 8e-5          & 0.1          & 1e-6         & 10                      \\
		TerraIncognita & 3e-5          & 0            & 1e-4         & 5e-6                    \\
		DomainNet      & 3e-5          & 0.1          & 0            & 5e-7                    \\ \bottomrule
	\end{tabular}
\end{table}

\subsection{Main results with variance}
In this chapter, we will provide specific experimental results for each combination of methods on FT-LP, as well as the main results containing variance.
\begin{table}[htbp]
	\centering
	\setlength{\tabcolsep}{3pt}
	\renewcommand{\arraystretch}{1.25}
	\caption{Main result with variance }
	\begin{tabular}{@{}lcccccc@{}}
		\bottomrule
		\toprule
		\multicolumn{1}{l|}{\textbf{Algorithm}} & {\texttt{PACS}}                              & {\texttt{VLCS}}                               & {\texttt{OfficeHome}}                          & {\texttt{TerraInc}}                              & \multicolumn{1}{c|}{\texttt{{DomainNet}} }                          & Avg.                 \\ \midrule
		\multicolumn{1}{l|}{$\mathrm{MMD}$\cite{MMD}}                & 84.7$\pm$\small 0.5 & 77.5$\pm$\small 0.9 & 66.3$\pm$\small 0.1 & 44.2$\pm$\small 1.6 & \multicolumn{1}{c|}{23.4$\pm$\small 9.5} & 58.8                 \\
		\multicolumn{1}{l|}{$\mathrm{IRM}$\cite{ref15}}                & 83.5$\pm$\small 0.8 & 78.5$\pm$\small 0.5 & 64.3$\pm$\small 2.2 & 47.6$\pm$\small 0.8 & \multicolumn{1}{c|}{33.9$\pm$\small 2.8} & 61.6                 \\
		\multicolumn{1}{l|}{$\mathrm{VREx}$\cite{ref16}}               & 84.9$\pm$\small 0.6 & 78.3$\pm$\small 0.2 & 66.4$\pm$\small 0.6 & 46.4$\pm$\small 0.6 & \multicolumn{1}{c|}{33.6$\pm$\small 2.9} & 61.9                 \\
		\multicolumn{1}{l|}{$\mathrm{CDANN}$\cite{CDANN}}              & 82.6$\pm$\small 0.9 & 77.5$\pm$\small 0.1 & 65.8$\pm$\small 1.3 & 45.8$\pm$\small 1.6 & \multicolumn{1}{c|}{38.3$\pm$\small 0.3} & 62.0                 \\
		\multicolumn{1}{l|}{$\mathrm{DANN}$\cite{DANN}}               & 83.6$\pm$\small 0.4 & 78.6$\pm$\small 0.4 & 65.9$\pm$\small 0.6 & 46.7$\pm$\small 0.5 & \multicolumn{1}{c|}{38.3$\pm$\small 0.1} & 62.6                 \\
		\multicolumn{1}{l|}{$\mathrm{Mixup}$\cite{Mixup}}              & 84.6$\pm$\small 0.6 & 77.4$\pm$\small 0.6 & 68.1$\pm$\small 0.3 & 47.9$\pm$\small 0.8 & \multicolumn{1}{c|}{39.2$\pm$\small 0.1} & 63.4                 \\
		\multicolumn{1}{l|}{$\mathrm{ERM}$\cite{domainbed}}                & 85.5$\pm$\small 0.2 & 77.3$\pm$\small 0.1 & 66.5$\pm$\small 0.2 & 46.1$\pm$\small 1.8 & \multicolumn{1}{c|}{43.8$\pm$\small 0.1} & 63.9                 \\
		\multicolumn{1}{l|}{$\mathrm{SagNet}$\cite{SagNet}}             & 86.3$\pm$\small 0.2 & 77.8$\pm$\small 0.5 & 68.1$\pm$\small 0.1 & 48.6$\pm$\small 1.0 & \multicolumn{1}{c|}{40.3$\pm$\small 0.1} & 64.2                 \\
		\multicolumn{1}{l|}{$\mathrm{RIDG}$\cite{RIDG}}             & 84.7$\pm$\small 0.2 & 77.8$\pm$\small 0.4 & 68.6$\pm$\small 0.2 & 47.8$\pm$\small 1.1 & \multicolumn{1}{c|}{41.9$\pm$\small 0.3} & 64.2                 \\
		\multicolumn{1}{l|}{$\mathrm{CORAL}$\cite{CORAL}}              & 86.2$\pm$\small 0.3 & 78.8$\pm$\small 0.6 & 68.7$\pm$\small 0.3 & 47.6$\pm$\small 1.0 & \multicolumn{1}{c|}{41.5$\pm$\small 0.1} & 64.5                 \\
		\multicolumn{1}{l|}{$\mathrm{SAGM}$\cite{SAGM}}               & 86.4$\pm$\small 0.6       & 79.1$\pm$\small 0.6        & 69.4$\pm$\small 0.2 & 48.0$ \pm$\small 0.4        & \multicolumn{1}{c|}{44.8$\pm$\small0.2
		}         & 65.6                     \\ \midrule
		\multicolumn{7}{l}{\textit{Combined with} FT-LP}                                                                                                                                                                                                                             \\ \midrule
		\multicolumn{1}{l|}{$\mathrm{MMD}$}                & 84.7$\pm$\small 1.9 & 78.4$\pm$\small 0.5 & 68.4$\pm$\small 0.1 & 47.5$\pm$\small 1.1 & \multicolumn{1}{c|}{43.1$\pm$\small 0.5}         & \multicolumn{1}{l}{64.4} \\
		\multicolumn{1}{l|}{$\mathrm{IRM}$}                & 85.4$\pm$\small 0.4 & 78.2$\pm$\small 0.8 & 67.1$\pm$\small 0.3 & 48.6$\pm$\small 0.6 & \multicolumn{1}{c|}{41.9$\pm$\small 0.1} & \multicolumn{1}{l}{63.8} \\
		\multicolumn{1}{l|}{$\mathrm{VREx}$}               & 86.4$\pm$\small 1.0 & 78.5$\pm$\small 0.3 & 67.2$\pm$\small 0.1 & 47.3$\pm$\small 1.5 & \multicolumn{1}{c|}{41.8$\pm$\small 0.3}         & \multicolumn{1}{l}{64.2} \\
		\multicolumn{1}{l|}{$\mathrm{CDANN}$}              & 84.0$\pm$\small 1.6 & 77.9$\pm$\small 0.3 & 67.8$\pm$\small 0.2 & 48.2$\pm$\small 0.9 & \multicolumn{1}{c|}{38.5$\pm$\small 0.2} & \multicolumn{1}{l}{63.3} \\
		\multicolumn{1}{l|}{$\mathrm{DANN}$}               & 84.4$\pm$\small 1.1 & 79.0$\pm$\small 1.0 & 67.8$\pm$\small 0.2 & 49.0$\pm$\small 1.5 & \multicolumn{1}{c|}{38.6$\pm$\small 0.1} & \multicolumn{1}{l}{63.8} \\
		\multicolumn{1}{l|}{$\mathrm{Mixup}$}              & 84.8$\pm$\small 1.4 & 79.0$\pm$\small 0.8 & 68.2$\pm$\small 0.1 & 48.1$\pm$\small 0.5 & \multicolumn{1}{c|}{43.3$\pm$\small 0.3}         & \multicolumn{1}{l}{64.7} \\
		\multicolumn{1}{l|}{$\mathrm{ERM}$}                & 86.2$\pm$\small 0.9 & 78.8$\pm$\small 0.4 & 68.7$\pm$\small 0.3 & 48.2$\pm$\small 2.1 & \multicolumn{1}{c|}{43.4$\pm$\small 0.4}         & \multicolumn{1}{l}{65.0} \\
		\multicolumn{1}{l|}{$\mathrm{SagNet}$}             & 85.8$\pm$\small 0.2 & 78.8$\pm$\small 1.0 & 68.1$\pm$\small 0.2 & \textbf{50.0$\pm$\small 0.6} & \multicolumn{1}{c|}{43.4$\pm$\small0.4}         & \multicolumn{1}{l}{65.2} \\
		\multicolumn{1}{l|}{$\mathrm{RIDG}$}             & 85.5$\pm$\small 0.8 & 78.7$\pm$\small 0.8 & 68.5$\pm$\small 0.3 & 48.0$\pm$\small 1.3 & \multicolumn{1}{c|}{42.0$\pm$\small0.3}         & \multicolumn{1}{l}{64.5} \\
		\multicolumn{1}{l|}{$\mathrm{CORAL}$}              & 86.3$\pm$\small 0.8 & 79.0$\pm$\small 0.5 & 68.5$\pm$\small 0.1 & 48.5$\pm$\small 1.9        & \multicolumn{1}{c|}{44.3$\pm$\small0.0}         & \multicolumn{1}{l}{65.3} \\
		\multicolumn{1}{l|}{$\mathrm{SAGM}$}               & \textbf{86.7$\pm$\small 0.4} & \textbf{79.2$\pm$\small 0.3} & \textbf{69.4$\pm$\small 0.4} & 48.6$\pm$\small 2.0 & \multicolumn{1}{c|}{\textbf{44.9$\pm$\small 0.2}}         & \multicolumn{1}{l}{65.8} \\ \bottomrule	\bottomrule
	\end{tabular}
\end{table}
\clearpage
\subsection{The effect of encoder $G$}
In the main text, we introduce encoders $G$ to our optimization objective. We believe that using $G$ can better fit the pre-training prior distribution and enhance generalization. Here, we use ablation experiments to validate the rationale. The initial part of the table \ref{xiao} displays the results without incorporating our FT-LP method, followed by the results after fine-tuning with FT-LP. Subsequently, we fixed the covariance of the prior model to a simple diagonal matrix, where $\Sigma = 1$ represents fixing our prior covariance to a diagonal matrix with all diagonal values equal to 1 : $\Sigma = diag(0.1,...,0.1)$. We keep all other hyperparameters consistent with the combined FT-LP case and obtain the  ablation results.

\begin{table}[htbp]

	\centering
	\setlength{\tabcolsep}{8pt}
	\renewcommand{\arraystretch}{1.25}
	\caption{The effect of encoders $G$ experiments}
		\label{xiao}
	\begin{tabular}{@{}lllll@{}}
		
		\toprule
		Algorithm & PACS                  & VLCS                  & Office                & Avg. \\ \midrule
		ERM       & 85.5$\pm$\small 0.2   & 77.3$\pm$\small 0.1   & 66.5$\pm$\small 0.2   & 76.4 \\
		IRM       & 83.5$\pm$\small 0.8   & 78.5$\pm$\small 0.5   & 64.3$\pm$\small 2.2   & 75.4 \\
		RIDG      & 84.7$\pm$\small 0.2   & 77.8$\pm$\small 0.4   & 68.6$\pm$\small 0.2   & 77.0 \\ \midrule
		\multicolumn{5}{l}{Combined with FT-LP}                                                  \\ \midrule
		ERM       & \textbf{86.2}$\pm$\small 0.9   & 78.8$\pm$\small 0.4   &\textbf{ 68.7}$\pm$\small 0.3   & 77.9 \\
		IRM       & 85.4$\pm$\small   0.4 & 78.2$\pm$\small   0.8 & 67.1$\pm$\small   0.3 & 76.9 \\
		RIDG      & 85.5$\pm$\small   0.8 & 78.7$\pm$\small   0.8 & 68.5$\pm$\small   0.3 & 77.6 \\ \midrule
		\multicolumn{5}{l}{Fixed $\Sigma = 0.1$}                                                 \\ \midrule
		ERM       & 84.4$\pm$\small   1.2 & 78.5$\pm$\small   0.2 & 65.8$\pm$\small   0.2 & 76.2 \\
		IRM       & 84.4$\pm$\small   0.7 & 77.9$\pm$\small   0.2 & 66.1$\pm$\small   0.3 & 76.1 \\
		RIDG      & 84.2$\pm$\small   0.8 & \textbf{79.1}$\pm$\small   0.2 & 68.7$\pm$\small   0.2 & 77.5 \\ \midrule
		\multicolumn{5}{l}{Fixed $\Sigma = 1$}                                                   \\ \midrule
		ERM       & 84.5$\pm$\small   1.2 & 78.6$\pm$\small   1.0 & 64.3$\pm$\small   0.1 & 75.8 \\
		IRM       & 85.1$\pm$\small   0.1 & 77.8$\pm$\small   0.3 &  67.1$\pm$\small   0.3& 76.7 \\
		RIDG      & 85.5$\pm$\small   0.1 & 78.5$\pm$\small   1.8 & 68.4$\pm$\small   0.7 & 77.3 \\ \bottomrule
	\end{tabular}
\end{table}

It can be observed that after fixing the covariance, the accuracy of all methods declines to varying degrees. However, the extent of decline varies among different algorithms. ERM experiences the most significant decrease, even performing worse than the initial results. The other two methods also experience declines, but they still surpass the initial situation. This indicates that incorporating the pre-trained model as a prior distribution into the optimization objective is indeed beneficial. Nevertheless, the choice of covariance still significantly influences the results. Manual selection of covariance is challenging to optimize, whereas encoding through our structural approach can further improve the model's generalization performance.
\clearpage
\subsection{Model selection: training-domain validation set}

\begin{table}[htbp]
		\centering
	\setlength{\tabcolsep}{8pt}
\renewcommand{\arraystretch}{1.2}
	\caption{Details in PACS }
	\begin{tabular}{@{}l|llllll@{}}
		\toprule
		{\color[HTML]{000000} Algorithm} & 1      & 2      & 3      & Avg. & var   & std \\ \midrule
		MMD                              & 86.844 & 84.162 & 83.208 & 84.7 & 3.554 & 1.9 \\
		IRM                              & 85.255 & 85.031 & 85.804 & 85.4 & 0.15 & 0.4 \\
		VREX                             & 85.367 & 86.415 & 87.464 & 86.4 & 1.099 & 1.0   \\
		CDANN                            & 85.643 & 83.997 & 82.408 & 84.0  & 2.617 & 1.6 \\
		DANN                             & 85.563 & 84.25  & 83.467 & 84.4 & 1.122 & 1.1 \\
		MIXUP                            & 84.717 & 86.161 & 83.412 & 84.8 & 1.891 & 1.4 \\
		ERM                              & 86.325 & 87.12  & 85.183 & 86.2 & 0.948 & 1.0  \\
		SAGNET                           & 85.749 & 85.726 & 86.027 & 85.8 & 0.028 & 0.2 \\
		RIDG                             & 86.435 & 84.981 & 85.191 & 85.5 & 0.618 & 0.8 \\
		CORAL                            & 87.051 & 85.54  & 86.27  & 86.3 & 0.571 & 0.8 \\
		SAGM                             & 87.187 & 86.561 & 86.383 & 86.7 & 0.178 & 0.4 \\ \bottomrule	\bottomrule
	\end{tabular}
\end{table}


\begin{table}[htbp]
		\centering
	\setlength{\tabcolsep}{8pt}
	\renewcommand{\arraystretch}{1.2}
	\caption{Details in VLCS }
	\begin{tabular}{@{}l|llllll@{}}
		\toprule
		Algorithm & 1      & 2      & 3      & Avg. & var   & std \\ \midrule
		MMD       & 78.491 & 78.858 & 77.838 & 78.4 & 0.267 & 0.5 \\
		IRM       & 78.882 & 78.208 & 77.376 & 78.2 & 0.569 & 0.8 \\
		VREX      & 78.559 & 78.818 & 78.216 & 78.5 & 0.091 & 0.3 \\
		CDANN     & 77.82  & 78.293 & 77.725 & 77.9 & 0.093 & 0.3 \\
		DANN      & 78.339 & 78.463 & 80.082 & 79.0 & 0.946 & 1.0 \\
		MIXUP     & 79.172 & 79.643 & 78.082 & 79.0 & 0.641 & 0.8 \\
		ERM       & 79.263 & 78.792 & 78.438 & 78.8 & 0.171 & 0.4 \\
		SAGNET    & 79.127 & 79.499 & 77.678 & 78.8 & 0.926 & 1.0 \\
		RIDG      & 79.494 & 78.429 & 78.044 & 78.7 & 0.564 & 0.8 \\
		CORAL     & 79.464 & 79.025 & 78.459 & 79.0 & 0.254 & 0.5 \\
		SAGM      & 79.178 & 78.845 & 79.527 & 79.2 & 0.116 & 0.3 \\ \bottomrule
	\end{tabular}
\end{table}

\begin{table}[htbp]
	\centering
	\setlength{\tabcolsep}{8pt}
\renewcommand{\arraystretch}{1.2}
	\caption{Details in OfficeHome }
	\begin{tabular}{@{}l|llllll@{}}
		\toprule
		Algorithm & 1      & 2      & 3      & Avg. & var   & std \\ \midrule
		MMD       & 68.538 & 68.284 & 68.398 & 68.4 & 0.016 & 0.1 \\
		IRM       & 66.952 & 67.37 & 66.868 & 67.1 & 0.080 & 0.3 \\
		VREX      & 67.202 & 67.149 & 67.327 & 67.2 & 0.008 & 0.1 \\
		CDANN     & 67.878 & 67.593 & 67.933 & 67.8 & 0.033 & 0.2 \\
		DANN      & 67.907 & 68.011 & 67.599 & 67.8 & 0.046 & 0.2 \\
		MIXUP     & 68.188 & 68.105 & 68.2   & 68.2 & 0.003 & 0.1 \\
		ERM       & 69.04  & 68.56  & 68.534 & 68.7 & 0.081 & 0.3 \\
		SAGNET    & 68.008 & 68.048 & 68.3   & 68.1 & 0.025 & 0.2 \\
		RIDG      & 68.674 & 68.598 & 68.183 & 68.5 & 0.07  & 0.3 \\
		CORAL     & 68.568 & 68.384 & 68.488 & 68.5 & 0.009 & 0.1 \\
		SAGM      & 69.708 & 69.522 & 69.22  & 69.5 & 0.061 & 0.2 \\ \bottomrule
	\end{tabular}
\end{table}


\begin{table}[htbp]
		\centering
	\setlength{\tabcolsep}{8pt}
	\renewcommand{\arraystretch}{1.2}
		\caption{Details in TerrInc. }
	\begin{tabular}{@{}l|llllll@{}}
		\toprule
		Algorithm & 1      & 2      & 3      & Avg.                        & var   & std \\ \midrule
		MMD       & 47.627 & 48.563 & 46.336 & 47.5                        & 1.25  & 1.1 \\
		IRM       & 49.087 & 48.692 & 47.97  & 48.6                        & 0.321 & 0.6 \\
		VREX      & 47.129 & 48.803 & 45.87  & 47.3                        & 2.165 & 1.5 \\
		CDANN     & 48.991 & 47.22  & 48.25  & 48.2                        & 0.791 & 0.9 \\
		DANN      & 50.325 & 47.441 & 49.361 & 49.0                        & 2.156 & 1.5 \\
		MIXUP     & 48.293 & 48.452 & 47.462 & 48.1                        & 0.283 & 0.5 \\
		ERM       & 49.384 & 49.364 & 45.755 & 48.2                        & 4.366 & 2.1 \\
		SAGNET    & 50.68  & 49.736 & 49.558 & 50.0                        & 0.364 & 0.6 \\
		RIDG      & 48.105 & 49.172 & 46.634 &  47.9 & 1.624 & 1.3 \\
		CORAL     & 50.421 & 48.58  & 46.572 &  48.5 & 3.706 & 1.9 \\
		SAGM      & 50.981 & 48.144 & 46.572 & 48.6                        & 4.993 & 2.2 \\ \bottomrule
	\end{tabular}
\end{table}

\begin{table}[htbp]
			\centering
	\setlength{\tabcolsep}{8pt}
	\renewcommand{\arraystretch}{1.2}
	\caption{Details in Domainbed }
	\begin{tabular}{@{}lllllll@{}}
		\toprule
		Algorithm                   & 1      & 2      & 3      & Avg. & var   & std \\ \midrule
		\multicolumn{1}{l|}{MMD}    & 43.516 & 43.197 & 42.587 & 43.1 & 0.223 & 0.5 \\
		\multicolumn{1}{l|}{IRM}    & 42.01  & 41.904 & 41.87  & 41.9 & 0.005 & 0.1 \\
		\multicolumn{1}{l|}{VREX}   & 42.052 & 41.99  & 41.488 & 41.8 & 0.096 & 0.3 \\
		\multicolumn{1}{l|}{CDANN}  & 38.621 & 38.251 & 38.499 & 38.5 & 0.036 & 0.2 \\
		\multicolumn{1}{l|}{DANN}   & 38.629 & 38.521 & 38.548 & 38.6 & 0.003 & 0.1 \\
		\multicolumn{1}{l|}{MIXUP}  & 43.596 & 43.258 & 43.173 & 43.3 & 0.05  & 0.2 \\
		\multicolumn{1}{l|}{ERM}    & 43.733 & 43.452 & 42.882 & 43.4 & 0.188 & 0.4 \\
		\multicolumn{1}{l|}{SAGNET} & 43.064 & 43.699 & 43.435 & 43.4 & 0.102 & 0.3 \\
		\multicolumn{1}{l|}{RIDG}   & 42.354 & 41.773 & 41.748 & 42.0   & 0.118 & 0.3 \\
		\multicolumn{1}{l|}{CORAL}  & 44.318 & 44.29  & 44.366 & 44.3 & 0.001 & 0.0  \\
		\multicolumn{1}{l|}{SAGM}   & 45.079 & 44.907 & 44.736 & 44.9 & 0.029 & 0.2 \\ \bottomrule
	\end{tabular}
\end{table}
\clearpage
\subsection{More visualizations }
\begin{figure}[t]
	\centering
	\includegraphics[width=0.9\textwidth]{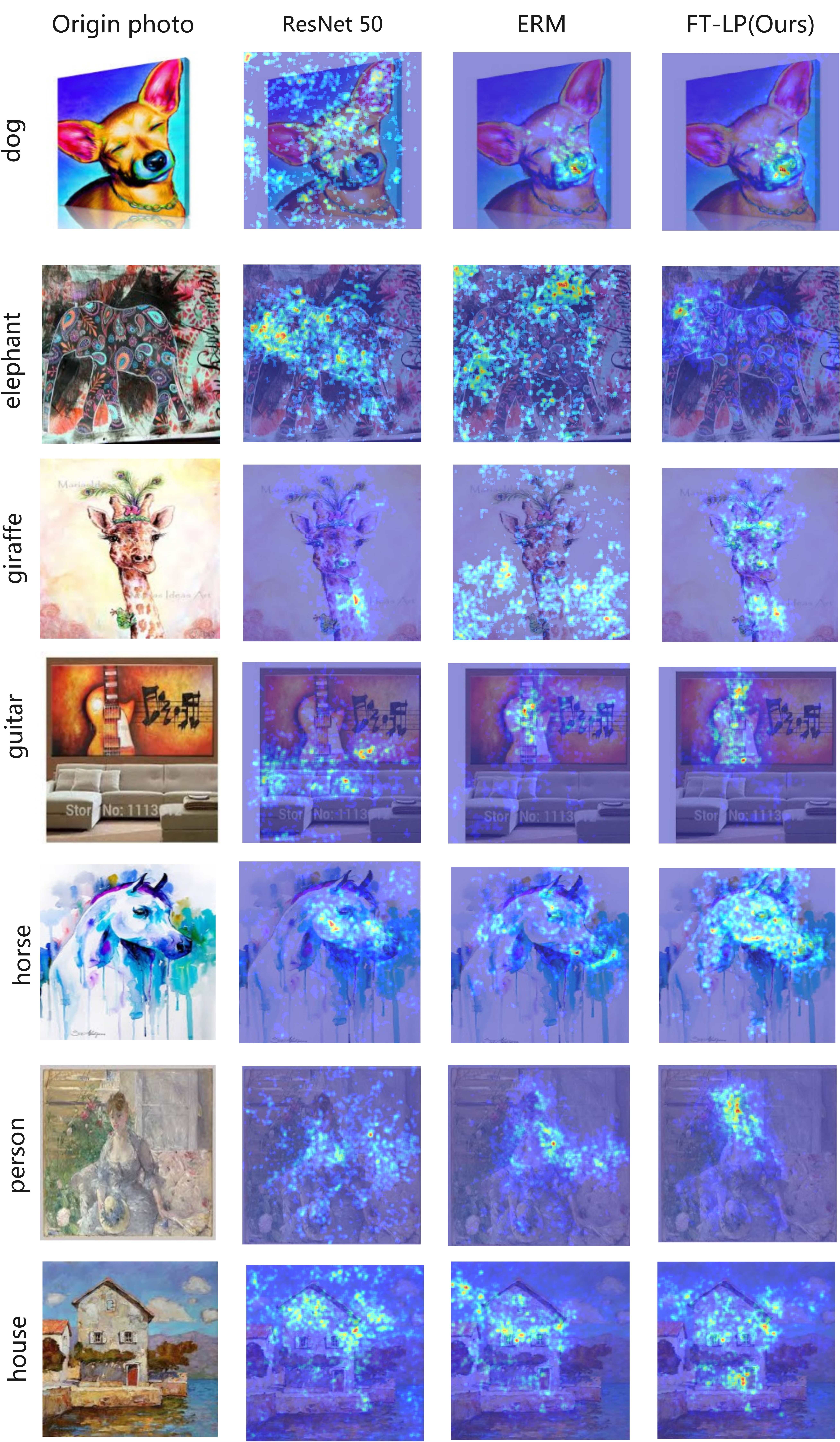}
	\caption{art painting}
	\label{fig4}
\end{figure}

\begin{figure}[t]
	\centering
	\includegraphics[width=0.9\textwidth]{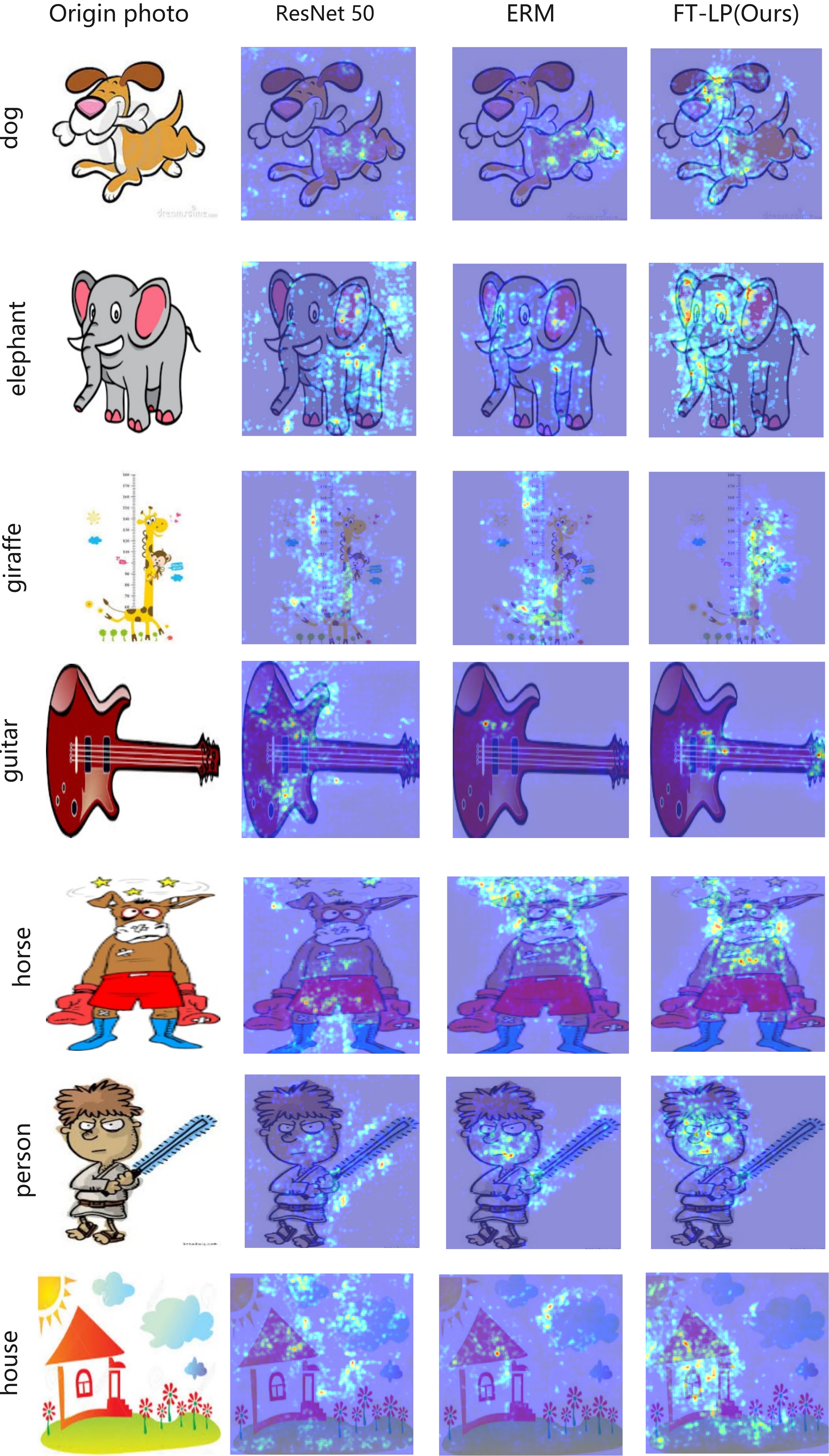}
	\caption{cartoon}
	\label{fig6}
\end{figure}

\begin{figure}[t]
	\centering
	\includegraphics[width=0.9\textwidth]{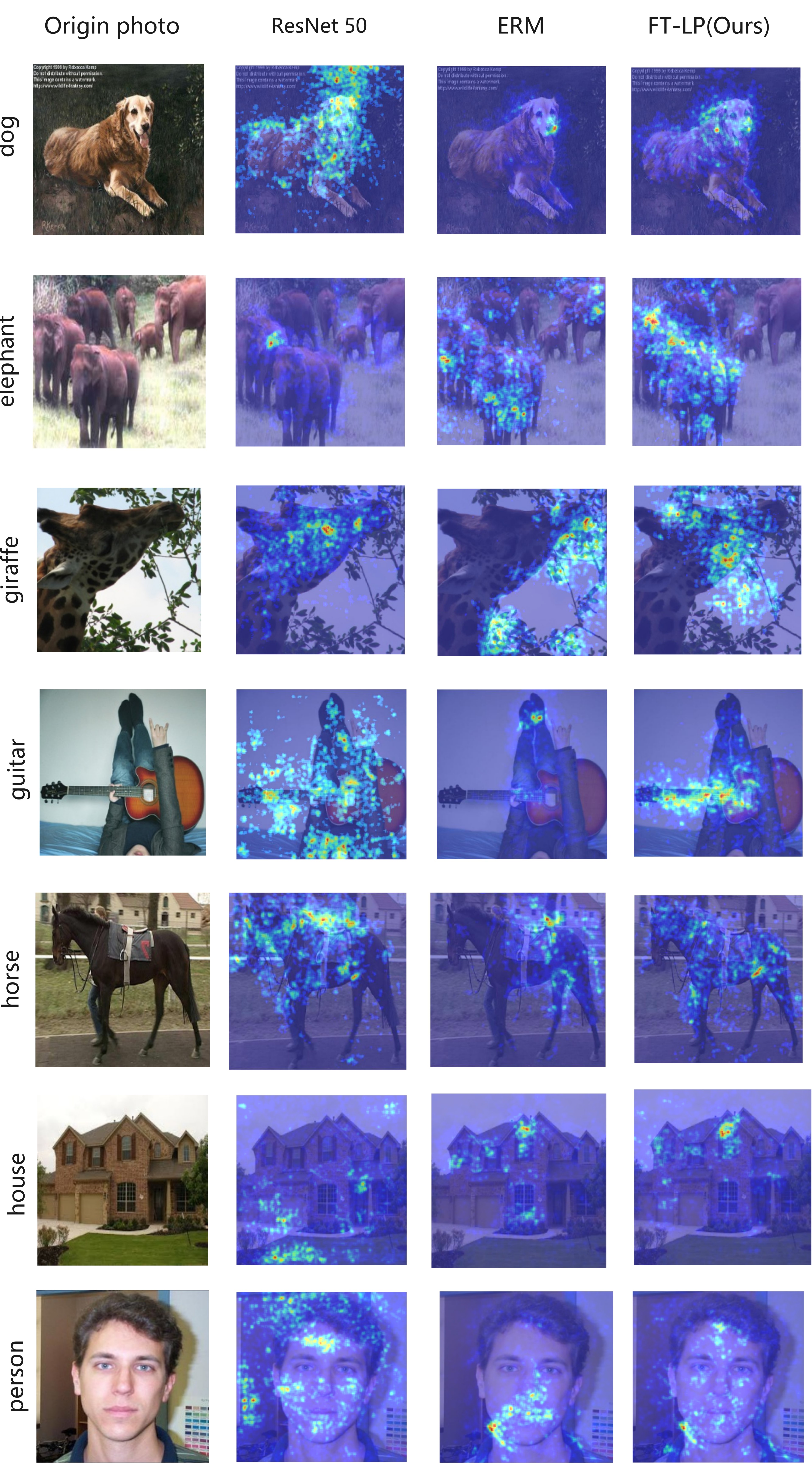}
	\caption{photo}
	\label{fig7}
\end{figure}

\begin{figure}[t]
	\centering
	\includegraphics[width=0.9\textwidth]{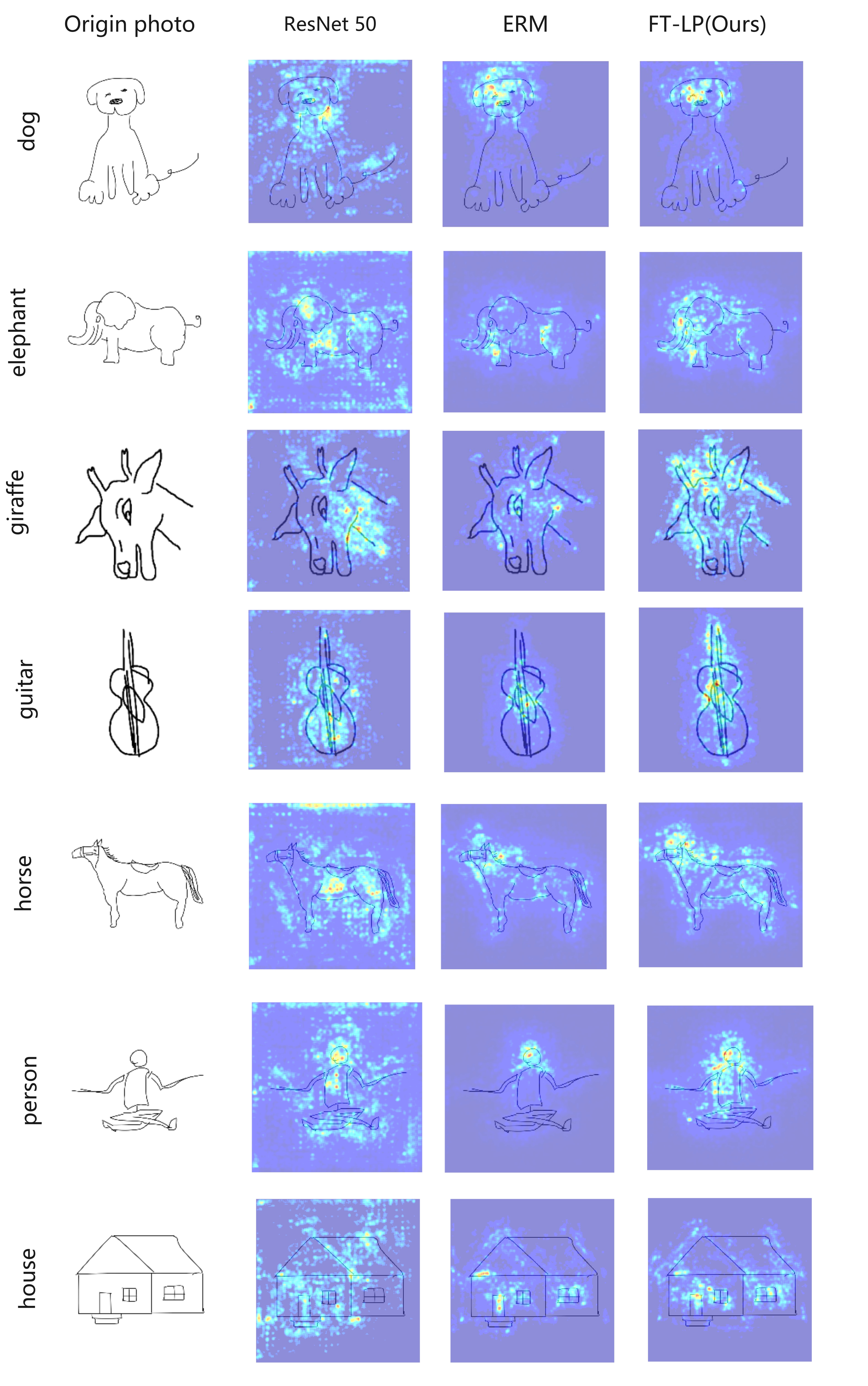}
	\caption{sketch}
	\label{fig8}
\end{figure}

In this chapter, we supplement more visualization images to validate our conclusions. The domain mentioned in the annotation of each image represents the target domain of that particular model. The model selection method we have chosen is consistent with the main results. As shown in Fig.\ref{fig4} to Fig.\ref{fig8}

\end{appendices}
\clearpage
\section*{NeurIPS Paper Checklist}

\begin{enumerate}
	
	\item {\bf Claims}
	\item[] Question: Do the main claims made in the abstract and introduction accurately reflect the paper's contributions and scope?
	\item[] Answer: \answerYes{} 
	\item[] Justification: 
	\item[] Guidelines:
	\begin{itemize}
		\item The answer NA means that the abstract and introduction do not include the claims made in the paper.
		\item The abstract and/or introduction should clearly state the claims made, including the contributions made in the paper and important assumptions and limitations. A No or NA answer to this question will not be perceived well by the reviewers. 
		\item The claims made should match theoretical and experimental results, and reflect how much the results can be expected to generalize to other settings. 
		\item It is fine to include aspirational goals as motivation as long as it is clear that these goals are not attained by the paper. 
	\end{itemize}
	
	\item {\bf Limitations}
	\item[] Question: Does the paper discuss the limitations of the work performed by the authors?
	\item[] Answer: \answerYes{} 
	\item[] Justification: It is in the \nameref{dissc}.
	\item[] Guidelines:
	\begin{itemize}
		\item The answer NA means that the paper has no limitation while the answer No means that the paper has limitations, but those are not discussed in the paper. 
		\item The authors are encouraged to create a separate "Limitations" section in their paper.
		\item The paper should point out any strong assumptions and how robust the results are to violations of these assumptions (e.g., independence assumptions, noiseless settings, model well-specification, asymptotic approximations only holding locally). The authors should reflect on how these assumptions might be violated in practice and what the implications would be.
		\item The authors should reflect on the scope of the claims made, e.g., if the approach was only tested on a few datasets or with a few runs. In general, empirical results often depend on implicit assumptions, which should be articulated.
		\item The authors should reflect on the factors that influence the performance of the approach. For example, a facial recognition algorithm may perform poorly when image resolution is low or images are taken in low lighting. Or a speech-to-text system might not be used reliably to provide closed captions for online lectures because it fails to handle technical jargon.
		\item The authors should discuss the computational efficiency of the proposed algorithms and how they scale with dataset size.
		\item If applicable, the authors should discuss possible limitations of their approach to address problems of privacy and fairness.
		\item While the authors might fear that complete honesty about limitations might be used by reviewers as grounds for rejection, a worse outcome might be that reviewers discover limitations that aren't acknowledged in the paper. The authors should use their best judgment and recognize that individual actions in favor of transparency play an important role in developing norms that preserve the integrity of the community. Reviewers will be specifically instructed to not penalize honesty concerning limitations.
	\end{itemize}
	
	\item {\bf Theory Assumptions and Proofs}
	\item[] Question: For each theoretical result, does the paper provide the full set of assumptions and a complete (and correct) proof?
	\item[] Answer: \answerYes{} 
	\item[] Justification: The proof for our propositions in in the Appendix \ref{proof}.
	\item[] Guidelines:
	\begin{itemize}
		\item The answer NA means that the paper does not include theoretical results. 
		\item All the theorems, formulas, and proofs in the paper should be numbered and cross-referenced.
		\item All assumptions should be clearly stated or referenced in the statement of any theorems.
		\item The proofs can either appear in the main paper or the supplemental material, but if they appear in the supplemental material, the authors are encouraged to provide a short proof sketch to provide intuition. 
		\item Inversely, any informal proof provided in the core of the paper should be complemented by formal proofs provided in appendix or supplemental material.
		\item Theorems and Lemmas that the proof relies upon should be properly referenced. 
	\end{itemize}
	
	\item {\bf Experimental Result Reproducibility}
	\item[] Question: Does the paper fully disclose all the information needed to reproduce the main experimental results of the paper to the extent that it affects the main claims and/or conclusions of the paper (regardless of whether the code and data are provided or not)?
	\item[] Answer: \answerYes{} 
	\item[] Justification: We provide more details in our Appendix \ref{exp} to reproduce the main experimental results.
	\item[] Guidelines:
	\begin{itemize}
		\item The answer NA means that the paper does not include experiments.
		\item If the paper includes experiments, a No answer to this question will not be perceived well by the reviewers: Making the paper reproducible is important, regardless of whether the code and data are provided or not.
		\item If the contribution is a dataset and/or model, the authors should describe the steps taken to make their results reproducible or verifiable. 
		\item Depending on the contribution, reproducibility can be accomplished in various ways. For example, if the contribution is a novel architecture, describing the architecture fully might suffice, or if the contribution is a specific model and empirical evaluation, it may be necessary to either make it possible for others to replicate the model with the same dataset, or provide access to the model. In general. releasing code and data is often one good way to accomplish this, but reproducibility can also be provided via detailed instructions for how to replicate the results, access to a hosted model (e.g., in the case of a large language model), releasing of a model checkpoint, or other means that are appropriate to the research performed.
		\item While NeurIPS does not require releasing code, the conference does require all submissions to provide some reasonable avenue for reproducibility, which may depend on the nature of the contribution. For example
		\begin{enumerate}
			\item If the contribution is primarily a new algorithm, the paper should make it clear how to reproduce that algorithm.
			\item If the contribution is primarily a new model architecture, the paper should describe the architecture clearly and fully.
			\item If the contribution is a new model (e.g., a large language model), then there should either be a way to access this model for reproducing the results or a way to reproduce the model (e.g., with an open-source dataset or instructions for how to construct the dataset).
			\item We recognize that reproducibility may be tricky in some cases, in which case authors are welcome to describe the particular way they provide for reproducibility. In the case of closed-source models, it may be that access to the model is limited in some way (e.g., to registered users), but it should be possible for other researchers to have some path to reproducing or verifying the results.
		\end{enumerate}
	\end{itemize}

	\item {\bf Open access to data and code}
	\item[] Question: Does the paper provide open access to the data and code, with sufficient instructions to faithfully reproduce the main experimental results, as described in supplemental material?
	\item[] Answer: \answerYes{} 
	\item[] Justification:  The code is described in supplemental material.
	\item[] Guidelines:
	\begin{itemize}
		\item The answer NA means that paper does not include experiments requiring code.
		\item Please see the NeurIPS code and data submission guidelines (\url{https://nips.cc/public/guides/CodeSubmissionPolicy}) for more details.
		\item While we encourage the release of code and data, we understand that this might not be possible, so “No” is an acceptable answer. Papers cannot be rejected simply for not including code, unless this is central to the contribution (e.g., for a new open-source benchmark).
		\item The instructions should contain the exact command and environment needed to run to reproduce the results. See the NeurIPS code and data submission guidelines (\url{https://nips.cc/public/guides/CodeSubmissionPolicy}) for more details.
		\item The authors should provide instructions on data access and preparation, including how to access the raw data, preprocessed data, intermediate data, and generated data, etc.
		\item The authors should provide scripts to reproduce all experimental results for the new proposed method and baselines. If only a subset of experiments are reproducible, they should state which ones are omitted from the script and why.
		\item At submission time, to preserve anonymity, the authors should release anonymized versions (if applicable).
		\item Providing as much information as possible in supplemental material (appended to the paper) is recommended, but including URLs to data and code is permitted.
	\end{itemize}

	\item {\bf Experimental Setting/Details}
	\item[] Question: Does the paper specify all the training and test details (e.g., data splits, hyperparameters, how they were chosen, type of optimizer, etc.) necessary to understand the results?
	\item[] Answer: \answerYes{} 
	\item[] Justification: The Experiment setups and implementation details specify these, we test our method using the Domainbed evaluation protocols.
	\item[] Guidelines:
	\begin{itemize}
		\item The answer NA means that the paper does not include experiments.
		\item The experimental setting should be presented in the core of the paper to a level of detail that is necessary to appreciate the results and make sense of them.
		\item The full details can be provided either with the code, in appendix, or as supplemental material.
	\end{itemize}
	
	\item {\bf Experiment Statistical Significance}
	\item[] Question: Does the paper report error bars suitably and correctly defined or other appropriate information about the statistical significance of the experiments?
	\item[] Answer: \answerYes{} 
	\item[] Justification: The variance of the main results and result of each experiment is in the Appendix \ref{exp}.
	\item[] Guidelines:
	\begin{itemize}
		\item The answer NA means that the paper does not include experiments.
		\item The authors should answer "Yes" if the results are accompanied by error bars, confidence intervals, or statistical significance tests, at least for the experiments that support the main claims of the paper.
		\item The factors of variability that the error bars are capturing should be clearly stated (for example, train/test split, initialization, random drawing of some parameter, or overall run with given experimental conditions).
		\item The method for calculating the error bars should be explained (closed form formula, call to a library function, bootstrap, etc.)
		\item The assumptions made should be given (e.g., Normally distributed errors).
		\item It should be clear whether the error bar is the standard deviation or the standard error of the mean.
		\item It is OK to report 1-sigma error bars, but one should state it. The authors should preferably report a 2-sigma error bar than state that they have a 96\% CI, if the hypothesis of Normality of errors is not verified.
		\item For asymmetric distributions, the authors should be careful not to show in tables or figures symmetric error bars that would yield results that are out of range (e.g. negative error rates).
		\item If error bars are reported in tables or plots, The authors should explain in the text how they were calculated and reference the corresponding figures or tables in the text.
	\end{itemize}
	
	\item {\bf Experiments Compute Resources}
	\item[] Question: For each experiment, does the paper provide sufficient information on the computer resources (type of compute workers, memory, time of execution) needed to reproduce the experiments?
	\item[] Answer: \answerYes{} 
	\item[] Justification: The environment and computer resources is reported in the Appendix \ref{exp}.
	\item[] Guidelines:
	\begin{itemize}
		\item The answer NA means that the paper does not include experiments.
		\item The paper should indicate the type of compute workers CPU or GPU, internal cluster, or cloud provider, including relevant memory and storage.
		\item The paper should provide the amount of compute required for each of the individual experimental runs as well as estimate the total compute. 
		\item The paper should disclose whether the full research project required more compute than the experiments reported in the paper (e.g., preliminary or failed experiments that didn't make it into the paper). 
	\end{itemize}
	
	\item {\bf Code Of Ethics}
	\item[] Question: Does the research conducted in the paper conform, in every respect, with the NeurIPS Code of Ethics \url{https://neurips.cc/public/EthicsGuidelines}?
	\item[] Answer: \answerYes{} 
	\item[] Justification: 
	\item[] Guidelines:
	\begin{itemize}
		\item The answer NA means that the authors have not reviewed the NeurIPS Code of Ethics.
		\item If the authors answer No, they should explain the special circumstances that require a deviation from the Code of Ethics.
		\item The authors should make sure to preserve anonymity (e.g., if there is a special consideration due to laws or regulations in their jurisdiction).
	\end{itemize}

	\item {\bf Broader Impacts}
	\item[] Question: Does the paper discuss both potential positive societal impacts and negative societal impacts of the work performed?
	\item[] Answer: \answerNA{} 
	\item[] Justification: There is no societal impact of our work.
	\item[] Guidelines:
	\begin{itemize}
		\item The answer NA means that there is no societal impact of the work performed.
		\item If the authors answer NA or No, they should explain why their work has no societal impact or why the paper does not address societal impact.
		\item Examples of negative societal impacts include potential malicious or unintended uses (e.g., disinformation, generating fake profiles, surveillance), fairness considerations (e.g., deployment of technologies that could make decisions that unfairly impact specific groups), privacy considerations, and security considerations.
		\item The conference expects that many papers will be foundational research and not tied to particular applications, let alone deployments. However, if there is a direct path to any negative applications, the authors should point it out. For example, it is legitimate to point out that an improvement in the quality of generative models could be used to generate deepfakes for disinformation. On the other hand, it is not needed to point out that a generic algorithm for optimizing neural networks could enable people to train models that generate Deepfakes faster.
		\item The authors should consider possible harms that could arise when the technology is being used as intended and functioning correctly, harms that could arise when the technology is being used as intended but gives incorrect results, and harms following from (intentional or unintentional) misuse of the technology.
		\item If there are negative societal impacts, the authors could also discuss possible mitigation strategies (e.g., gated release of models, providing defenses in addition to attacks, mechanisms for monitoring misuse, mechanisms to monitor how a system learns from feedback over time, improving the efficiency and accessibility of ML).
	\end{itemize}
	
	\item {\bf Safeguards}
	\item[] Question: Does the paper describe safeguards that have been put in place for responsible release of data or models that have a high risk for misuse (e.g., pretrained language models, image generators, or scraped datasets)?
	\item[] Answer:  \answerNA{} 
	\item[] Justification:  Our paper poses no such risks.
	\item[] Guidelines:
	\begin{itemize}
		\item The answer NA means that the paper poses no such risks.
		\item Released models that have a high risk for misuse or dual-use should be released with necessary safeguards to allow for controlled use of the model, for example by requiring that users adhere to usage guidelines or restrictions to access the model or implementing safety filters. 
		\item Datasets that have been scraped from the Internet could pose safety risks. The authors should describe how they avoided releasing unsafe images.
		\item We recognize that providing effective safeguards is challenging, and many papers do not require this, but we encourage authors to take this into account and make a best faith effort.
	\end{itemize}
	
	\item {\bf Licenses for existing assets}
	\item[] Question: Are the creators or original owners of assets (e.g., code, data, models), used in the paper, properly credited and are the license and terms of use explicitly mentioned and properly respected?
	\item[] Answer: \answerYes{} 
	\item[] Justification: We listed original owners of assets in the section Experiment setups and implementation details.
	\item[] Guidelines:
	\begin{itemize}
		\item The answer NA means that the paper does not use existing assets.
		\item The authors should cite the original paper that produced the code package or dataset.
		\item The authors should state which version of the asset is used and, if possible, include a URL.
		\item The name of the license (e.g., CC-BY 4.0) should be included for each asset.
		\item For scraped data from a particular source (e.g., website), the copyright and terms of service of that source should be provided.
		\item If assets are released, the license, copyright information, and terms of use in the package should be provided. For popular datasets, \url{paperswithcode.com/datasets} has curated licenses for some datasets. Their licensing guide can help determine the license of a dataset.
		\item For existing datasets that are re-packaged, both the original license and the license of the derived asset (if it has changed) should be provided.
		\item If this information is not available online, the authors are encouraged to reach out to the asset's creators.
	\end{itemize}
	
	\item {\bf New Assets}
	\item[] Question: Are new assets introduced in the paper well documented and is the documentation provided alongside the assets?
	\item[] Answer: \answerNA{} 
	\item[] Justification: We do not release new assets, we are using the Domainbed evaluation protocols.
	\item[] Guidelines:
	\begin{itemize}
		\item The answer NA means that the paper does not release new assets.
		\item Researchers should communicate the details of the dataset/code/model as part of their submissions via structured templates. This includes details about training, license, limitations, etc. 
		\item The paper should discuss whether and how consent was obtained from people whose asset is used.
		\item At submission time, remember to anonymize your assets (if applicable). You can either create an anonymized URL or include an anonymized zip file.
	\end{itemize}
	
	\item {\bf Crowdsourcing and Research with Human Subjects}
	\item[] Question: For crowdsourcing experiments and research with human subjects, does the paper include the full text of instructions given to participants and screenshots, if applicable, as well as details about compensation (if any)? 
	\item[] Answer:  \answerNA{} 
	\item[] Justification: Our paper does not involve crowdsourcing nor research with human subjects.
	\item[] Guidelines:
	\begin{itemize}
		\item The answer NA means that the paper does not involve crowdsourcing nor research with human subjects.
		\item Including this information in the supplemental material is fine, but if the main contribution of the paper involves human subjects, then as much detail as possible should be included in the main paper. 
		\item According to the NeurIPS Code of Ethics, workers involved in data collection, curation, or other labor should be paid at least the minimum wage in the country of the data collector. 
	\end{itemize}
	
	\item {\bf Institutional Review Board (IRB) Approvals or Equivalent for Research with Human Subjects}
	\item[] Question: Does the paper describe potential risks incurred by study participants, whether such risks were disclosed to the subjects, and whether Institutional Review Board (IRB) approvals (or an equivalent approval/review based on the requirements of your country or institution) were obtained?
	\item[] Answer:  \answerNA{} 
	\item[] Justification: Our paper does not involve crowdsourcing nor research with human subjects.
	\item[] Guidelines:  
	\begin{itemize}
		\item The answer NA means that the paper does not involve crowdsourcing nor research with human subjects.
		\item Depending on the country in which research is conducted, IRB approval (or equivalent) may be required for any human subjects research. If you obtained IRB approval, you should clearly state this in the paper. 
		\item We recognize that the procedures for this may vary significantly between institutions and locations, and we expect authors to adhere to the NeurIPS Code of Ethics and the guidelines for their institution. 
		\item For initial submissions, do not include any information that would break anonymity (if applicable), such as the institution conducting the review.
	\end{itemize}
	
\end{enumerate}

\end{document}